\documentclass{ieeeaccess}
\usepackage{cite}
\usepackage{tabularx}
\usepackage{multirow}
\usepackage{graphicx}
\usepackage{float}
\usepackage{amsmath,amssymb,amsfonts }
\usepackage[colorlinks=acc.true,citecolor=blue,linkcolor=blue]{hyperref}%

\usepackage[table]{xcolor}

\usepackage{xcolor}
\usepackage{pict2e}

\newsavebox{\ORCIDlogo}
\savebox{\ORCIDlogo}{%
\setlength{\unitlength}{\dimexpr 1em/256\relax}%
\begin{picture}(256,256)%
  \color[HTML]{A6CE39}\put(128,128){\circle*{256}}%
  \color{white}%
  \put(78.6,199.2){\circle*{20}}%
  \moveto(70.9,176,9)\lineto(86.3,176,9)\lineto(86.3,69.8)\lineto(70.9,69.8)%
  \closepath\fillpath%
  \moveto(108.9,176.9)\lineto(150.5,176.9)%
  \curveto(190.1,176.9)(207.5,148.6)(207.5 ,123.3)%
  \curveto(207.5,95,8)(186,69.7)(150.7,69.7)%
  \lineto(108.9,69.7)%
  \closepath\fillpath%
  \color[HTML]{A6CE39}%
  \moveto(124.3,83.6)\lineto(148.8,83.6)%
  \curveto(183.7,83.6)(191.7,110.1)(191.7,123.3)%
  \curveto(191.7,144.8)(178,163)(148,163)%
  \lineto(124.3,163)%
  \closepath\fillpath%
\end{picture}%
}
\newcommand\orcidicon[1]{\href{https://orcid.org/#1}{\usebox{\ORCIDlogo}}}

\usepackage{algorithmic}

\usepackage{textcomp}
\usepackage[capitalise]{cleveref } 
\usepackage{siunitx}
\usepackage{float}
\usepackage[numbers,sort&compress]{natbib}
 
\def\textbf#1{{\bfseries #1}}
    
\def\BibTeX{{\rm B\kern-.05em{\sc i\kern-.025em b}\kern-.08em
    T\kern-.1667em\lower.7ex\hbox{E}\kern-.125emX}}

\begin{document}
\history{ }
\doi{ }

\title{A Cloud-based Deep Learning Framework for
Early Detection of Pushing at Crowded Event Entrances}
\author{\uppercase{Ahmed Alia\orcidicon{0000-0002-3049-4924}}\authorrefmark{1,2,3},  
\uppercase{Mohammed Maree\orcidicon{0000-0002-6114-4687}}\authorrefmark{4},
\uppercase{Mohcine Chraibi}\authorrefmark{1},
\uppercase{Anas Toma}\authorrefmark{5}, 
\uppercase{and Armin Seyfried\orcidicon{0000-0001-8888-0978}}\authorrefmark{1,2}, 
}

\address[1]{Institute for Advanced Simulation, Forschungszentrum Jülich, 52425 Jülich, Germany}
\address[2]{Department of Computer Simulation for Fire Protection and Pedestrian Traffic, University of Wuppertal,  42285  Wuppertal, Germany}
\address[3]{Department of Information Technology,  An-Najah National University, Nablus, Palestine}
\address[4]{Department of Information Technology, Arab American University, 00970 Jenin, Palestine}
\address[5]{Department of Electrical and Computer Engineering,  An-Najah National University, Nablus, Palestine}

\tfootnote{This work was funded by the German Federal Ministry of Education and Research (BMBF:
funding number 01DH16027) within the Palestinian-German Science Bridge project framework, and
partially by the Deutsche Forschungsgemeinschaft (DFG, German Research Foundation)—491111487. }

\markboth
{    }
{  }

\corresp{Corresponding authors: Ahmed Alia (a.alia@fz-juelich.de), Mohammed Maree (mohammed.maree@aaup.edu) and Armin Seyfried (a.seyfried@fz-juelich.de)}

\begin{abstract}
 
Crowding at the entrances of large events may lead to critical and life-threatening situations, particularly when people start pushing each other to reach the event faster. Automatic and timely identification of pushing behavior would help organizers and security forces to intervene early and mitigate dangerous situations. In this paper, we propose a cloud-based deep learning framework for automatic early detection of pushing in crowded event entrances. The proposed framework initially modifies and trains the EfficientNetV2B0 Convolutional Neural Network model. Subsequently, it integrates the adapted model with an accurate and fast pre-trained deep optical flow model with the color wheel method to analyze video streams and identify pushing patches in real-time. Moreover, the framework uses live capturing technology and a cloud-based environment to collect video streams of crowds in real-time and provide early-stage results. A novel dataset is generated based on five real-world experiments and their associated ground truth data to train the adapted EfficientNetV2B0 model. The experimental setups simulated a crowded event entrance, while the ground truths for each video experiment was generated manually by social psychologists. Several experiments on the videos and the generated dataset are carried out to evaluate the accuracy and annotation delay time of the proposed framework. The experimental results show that the proposed framework identified pushing behaviors with an accuracy rate of \qty{87}{\percent} within a reasonable delay time.
\end{abstract}


\titlepgskip=-21pt

\maketitle

\section{Introduction}
\label{sec:introduction}

The entrances of large-scale events such as sport venues, concerts, and religious gatherings are organized as bottlenecks for access control, ticket validation, or security check~\cite{adrian2020crowds}.
In these scenarios, some pedestrians might start pushing each other to gain faster access to the event.
According to Lügering et al.~\cite{usten2022pushing}, pushing for forward motion is defined as ``a behavior that can involve using arms, shoulders, or elbows; or simply the upper body, in which one person actively applies force to another person (or people) to overtake, while shifting their direction to the side or back, or force them to move forward more quickly.''
Additionally, using gaps in the crowd is considered as a strategy of pushing because it is a form of overtaking~\cite{usten2022pushing}. 
Indeed, such behavior increases the crowd's density over time~\cite{adrian2020crowds,haghani2019push}, resulting in the lack of comfort zones and, more importantly, can lead to dangerous situations~\cite{filingeri2017factors,johnson1987panic}.
In such cases, early pushing detection is essential, as it can provide valuable information to the organizers and the security team for better crowd management, thereby ensuring a smoother flow at entrances with higher safety~\cite{tyagi2022review}. 
Since manual identification of pushing behavior in the early stages can be complex or impossible, developing an automatic detection framework in real-time or near real-time is crucial.
However, automatic pushing detection is still a challenging task due to the highly-dense crowds, the diversity of pushing behavior strategies, and the varying features for pushing behavior representation, which still requires further investigation and identification~\cite{alia2022hybrid}.

Surveillance cameras have recently been widely integrated with computer vision techniques to automatically identify abnormal behaviors from crowds ~\cite{mehmood2021efficient,al2020transfer}. Within the realm of computer vision, pushing behavior can be classified as abnormal behavior. 
Machine learning algorithms, particularly Convolutional Neural Network (CNN) architectures, have remarkably succeeded in several computer vision tasks; among these is abnormal behavior detection in crowds~\cite{direkoglu2020abnormal}.
One of the critical reasons for this success is that CNN can learn the relevant features~\cite{alia2021enhanced, alia2017feature} and classification automatically from data without human intervention~\cite{gan2022spatiotemporal,gan2021automated}. 
Although CNN architectures are powerful for modeling human behaviors, building an accurate model requires a large training dataset~\cite{li2021apple,wang2021comparative}, which is often unavailable. Researchers have developed hybrid-based approaches that integrate CNN with handcrafted feature descriptors to address this limitation~\cite{duman2019anomaly,alafif2023hybrid}. These approaches employ descriptors to obtain useful data, which is subsequently used by CNN to learn and identify abnormal behavior automatically.
Due to the limited availability of labeled data for pushing behavior, hybrid-based approaches may be more appropriate for automatically identifying pushing behavior. For example, Alia et al.~\cite{alia2022hybrid} proposed a hybrid deep learning and visualization framework for pushing behavior detection in video recordings of crowded event entrances. Unfortunately, this framework does not cope with early detection requirements because it is slow and can not work with the live camera stream. To the best of our knowledge, despite the numerous computer vision and machine learning approaches reported in the literature, none of them can detect pushing behavior in real-time or near real-time from crowds. 

In order to address the above limitations, this article introduces a novel cloud-based deep learning framework for pushing patch detection in live video streams acquired from crowded event entrances. In this framework, we propose: 1) Integrating a robust deep optical flow model (GPU-based pre-trained Recurrent All-pairs Field Transforms (RAFT)~\cite{teed2020raft}) with the color wheel method~\cite{flow_vis,baker2011database} to accurately and rapidly extract the visual motion information from the crowd. 2) Adapting and training EfficientNetV2B0-based CNN~\cite{tan2021efficientnetv2} using visual motion information to detect pushing patches accurately. 3) Using live camera technology and a cloud environment to provide more powerful computational resources and help to collect and annotate the video stream of the crowd in real-time.

The main contributions of this article are summarized as follows:

\begin{enumerate}
\item To the best of our knowledge, we propose the first real-time or near real-time automatic framework dedicated to early identifying pushing behavior in human crowds.

\item We introduce a new video analysis and pushing detection approach based on integrating an adapted version of EfficentNetV2B0, GPU-based pre-trained RAFT model, and color wheel method.
  
\item  We create a novel dataset for pushing behavior, using five real-world experiments with their associated ground truths. This dataset is not only used as a training and evaluation resource for our adapted EfficientNetV2B0, but can also be a valuable asset for future research in this area.

\item We perform a thorough performance comparison of fifteen CNN architectures for pushing detection using the generated dataset.
 
\end{enumerate}

The rest of the paper is organized as follows.~\cref{sec:relatedwork}  reviews the related studies of video-based abnormal human behavior detection. The proposed framework is presented in~\cref{sec:propsedframework}. \cref{sec:evaluationandresults} discusses the evaluation process and experimental results.  Finally, the conclusion and future work are summarized in \cref{sec:conclusion}.

\section {Related Work}
\label{sec:relatedwork}

Generally, identifying pushing behavior in videos falls under the field of computer vision, specifically in the task of abnormal behavior detection. CNNs have played a crucial role in significant advancements in this area~\cite{kuppusamy2022human}. Consequently, in this section, our objective is to examine several abnormal behavior detection approaches that have been developed using CNNs.

A customized CNN-based method to identify abnormal activities in videos was presented by Tay et al~\cite{tay2019robust}.
The authors trained a customized CNN for feature extraction and labeling using normal and abnormal samples.
In another study, Alafif et al.~\cite{alafif2023hybrid} proposed two methods of identifying abnormal behaviors in small and large-scale crowd videos.
The first method employs a combination of a CNN model and a random forest classifier to detect anomaly behaviors at the object level in a small-scale crowd.
In contrast, the second method utilizes two classifiers to recognize abnormal behaviors in a large-scale crowd. 
The initial model, finds the frames containing abnormal behaviors, while the second classifier, You Only Look Once (version 2), processes those frames to identify abnormal behaviors exhibited by individuals. 
The effectiveness of these techniques relies heavily on utilizing CNNs to learn features from labeled datasets containing both normal and abnormal behaviors. A large training dataset of normal and abnormal behaviors is necessary to create an accurate and adaptable CNN model. However, obtaining such a dataset is often unattainable for various abnormal behaviors, including pushing behavior.

In order to overcome the shortage of large datasets comprising normal and abnormal behaviors, some researchers have utilized one-class classifiers with datasets consisting only of normal behaviors.
It is easier to obtain or create a dataset that contains only normal behavior than a dataset that includes both normal and abnormal behaviors~\cite {sabokrou2018deep,xu2019efficient}.
The fundamental concept behind the one-class classifier is to exclusively learn from normal behaviors, thereby establishing a class boundary between normal and undefined (abnormal) classes.
For example,
Sabokrou~et~al.~\cite {sabokrou2018deep}  employed a pre-trained CNN for extracting motion and appearance information from crowded scenes.
Subsequently, they utilized a one-class Gaussian distribution to construct the classifier using datasets comprised of normal behavior.
Similarly, in~\cite{xu2019efficient,smeureanu2017deep}, the authors developed one-class classifiers by utilizing a dataset of normal samples.
In~\cite{xu2019efficient}, Xu et al. employed a convolutional variational autoencoder to extract features, followed by the use of multiple Gaussian models to detect abnormal behavior.
Meanwhile, in~\cite{smeureanu2017deep}, a pre-trained CNN model was utilized for feature extraction, with one-class support vector machines being used to identify abnormal behavior.
Another study by Ilyas et al. \cite{ilyas2021hybrid} utilized a pre-trained CNN and a gradient sum of the frame difference to extract significant features.
Following this, three support vector machines were trained on normal behavior to detect abnormal behaviors. Generally, the one-class classifier is commonly used when the target behavior class or abnormal behavior is infrequent or poorly defined \cite{khan2014one}.
However, pushing behavior is well-defined and not rare, particularly in high-density and competitive situations. Furthermore, this type of classifier regards new normal behavior as abnormal.

To overcome the limitations of CNN-based and one-class classifier approaches, several studies have combined multi-class CNN with one or more handcrafted feature descriptors~\cite {ilyas2021hybrid,direkoglu2020abnormal}.
As an example, Duman et al.~\cite{duman2019anomaly} utilized the traditional  Farneb\"{a}ck optical flow approach in conjunction with CNN to detect anomalous behavior. They extracted direction and speed information using  Farneb\"{a}ck and CNN, and then utilized a convolutional long short-term memory network to construct the classifier.
Similarly, Hu et al.~\cite{hu2020design} employed a combination of the histogram of gradient and CNN for feature extraction, while a least-squares support vector was used for classification.
Almazroey et al.~\cite{almazroey2020abnormal}  focused on utilizing the Lucas-Kanade optical flow method, pre-trained CNN, and feature selection method (neighborhood component analysis) to extract relevant features. They then used a support vector machine to generate a trained classifier. In a different study~\cite{zhou2016spatial}, Zhou et al. introduced a CNN-based method to identify and locate abnormal activities. This approach integrated optical flow with CNN for feature extraction and utilized a CNN for classification.
Direkoglu~\cite{direkoglu2020abnormal} utilized the Lucas-Kanade optical flow method and CNN to extract relevant features and identify "escape and panic behaviors".

Most of the hybrid-based approaches for abnormal behavior detection that were reviewed have limited efficiency in detecting pushing since 1) The descriptors used in these approaches can only extract limited essential data from high-density crowds to represent pushing behavior.
2) Some CNN architectures commonly utilized in these approaches may not be effective in dealing with the increased variations within pushing behavior (intra-class variance) and the substantial resemblance between pushing and non-pushing behaviors (high inter-class similarity), which can potentially result in misclassification. To benefit from the power of hybrid-based approaches on a small dataset, Alia et al.~\cite{alia2022hybrid} introduced a hybrid framework for pushing patch detection in video recordings of crowds. The authors utilized a robust handcrafted feature descriptor and efficient CNN architecture in this framework. In more details, the framework used a deep optical flow technique to extract the motion information from the crowds. This information is then analyzed using an EfficientNetB0-based CNN and false reduction algorithms to identify and label pushing patches in the video.  However, this framework does not cope with early detection requirements due to three reasons.
First, it can only handle offline-recorded videos. Second, The deep optical flow technique employed in motion extraction is slow because it was performed on the CPU. Third,  it needs to identify pushing patches for the whole video before producing the output. 
Moreover, as reported by the authors, the accuracy of the framework decreases with complex scenarios of pushing.

\label{sec:propsedframework}

\begin{figure*}[ht]
\centering
\includegraphics[width=0.9\linewidth]{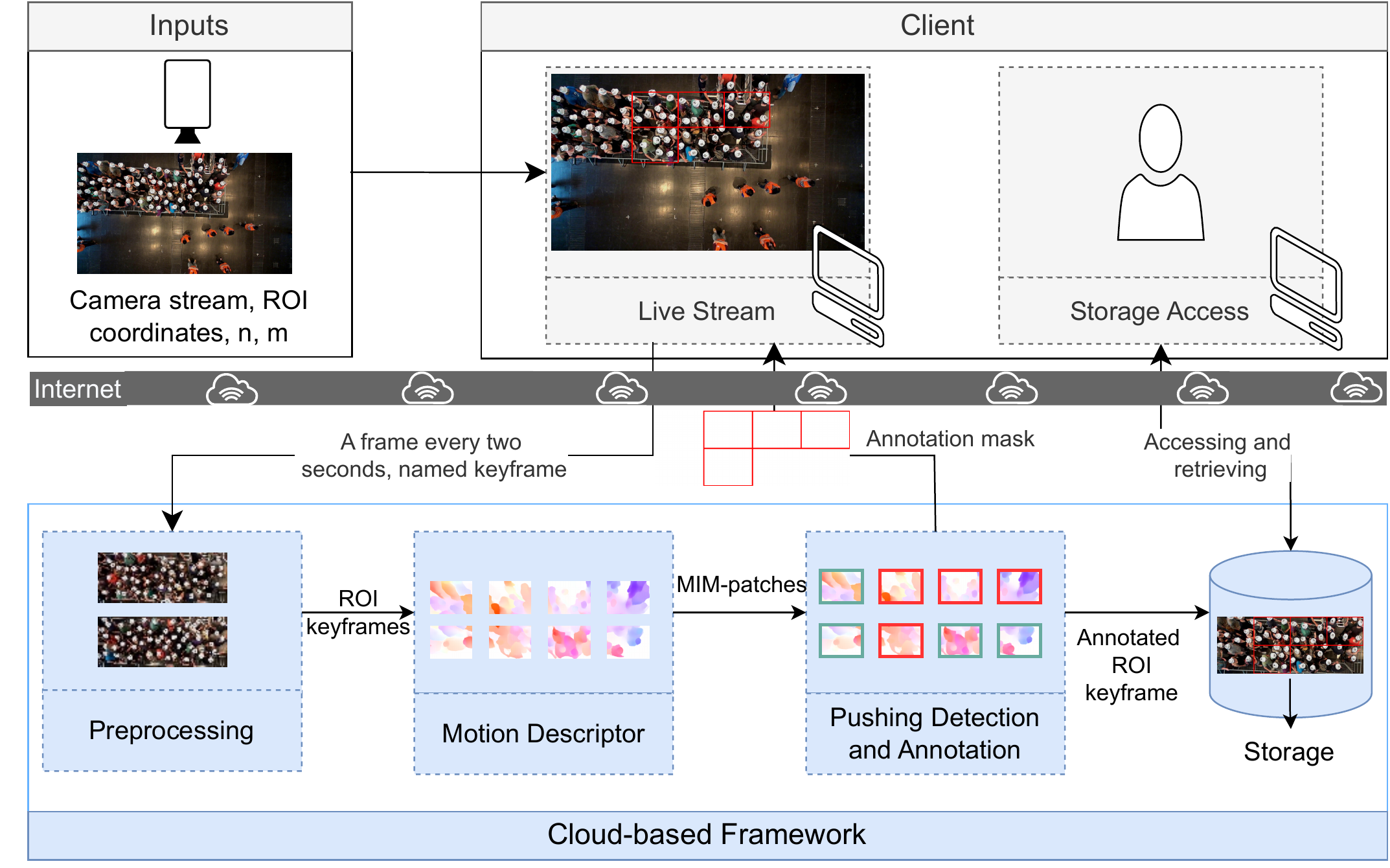}

\caption{The proposed framework architecture. ROI refers to the entrance area (Region Of Interest). User-defined row ($n$) and column ($m$) are used to split Motion Information Map (MIM) into  $n \times  m$ patches.
}
\label{fig:sys-arch}
\end{figure*}

To sum up, the reviewed methods have limitations regarding early pushing detection in crowded human environments. On the one hand, approaches that rely solely on CNNs for feature extraction require a large dataset containing normal and abnormal behaviors, which is typically unavailable for pushing scenarios. On the other hand,  one-class classifiers are often used for infrequent or poorly defined target behavior or abnormal behavior. However, pushing behavior is well-defined and common, particularly in high-density and competitive scenarios. Additionally, this type of classifier may misclassify new normal behavior as abnormal. Although hybrid-based approaches may be more suitable for pushing behavior, existing methods do not meet the requirements for early pushing detection in human crowds. To overcome these limitations, this article proposes a novel framework that adapts the EfficientNetV2B0 model and integrates it with GPU-based RAFT, wheel color method and live camera technology on a cloud platform. The following section provides a detailed discussion of the framework.

\section{The Proposed Framework}
\label{sec:propsedframework}

In this section, we describe the proposed framework for early detection of pushing within the live camera stream of crowded event entrances, where the camera is fixed and top-view.
~\cref{fig:sys-arch} shows the architecture of our framework which comprises three major components: preprocessing; motion descriptor; and pushing detection and annotation. The first component aims to collect and process the live camera stream, as well as display the stream on the web client in real-time. Simultaneously, the second component, the motion descriptor, employs the GPU-based RAFT model and color wheel method~\cite{flow_vis,baker2011database} to extract the visual motion information from the crowd.  Finally, the pushing detection and annotation component utilizes the adapted and trained EfficientNetV2B0 model to analyze the visual motion information and detect pushing patches. Notably, it directly annotates the regions that contain pushing behavior on the live stream on the web client. The following sections provide a more detailed discussion of the three components.

\subsection{Preprocessing } 

In order to reduce the computational time of the framework without sacrificing performance, the preprocessing component directly displays the client camera stream on the web client. At the same time, it collects only the data required for detection purposes from the live stream.
Let  $\{f^t\}$ represents the live camera stream, where $t$ is the time of the frame $f$ in the stream. Firstly, this component displays the live stream on the web client in real-time without uploading it to the cloud. Then,  a frame $f^t$ is collected from the stream every two seconds, hereafter referred to as keyframe (examples in~\cref{fig:examples}a). 
After that, this component utilizes the user-defined coordinates in pixel units to crop the entrance area $\bar{f^t}$ (ROI keyframe) from its corresponding keyframe $f^t$.
Finally, $\bar{f^t}$ is submitted as an input to the second component.
For the brevity, we name the ROI keyframe sequence $\bigl\{\bar{f}^t,\bar{f}^{t+ 2}, \bar{f}^{t+ 4},  \dots \bigr\}$ as 
 $\{ \bar{f}_i \, | \, i=1,2,3, \dots   \}$, where $i$ is the order of the ROI keyframe in the stream, and $t$ is the time in seconds. 
~\cref{fig:examples}b displays two examples of $\bar{f_i}$.

\subsection{Motion descriptor} 

\begin{figure*}[ht] 
\centering
\includegraphics[width=0.9\linewidth]{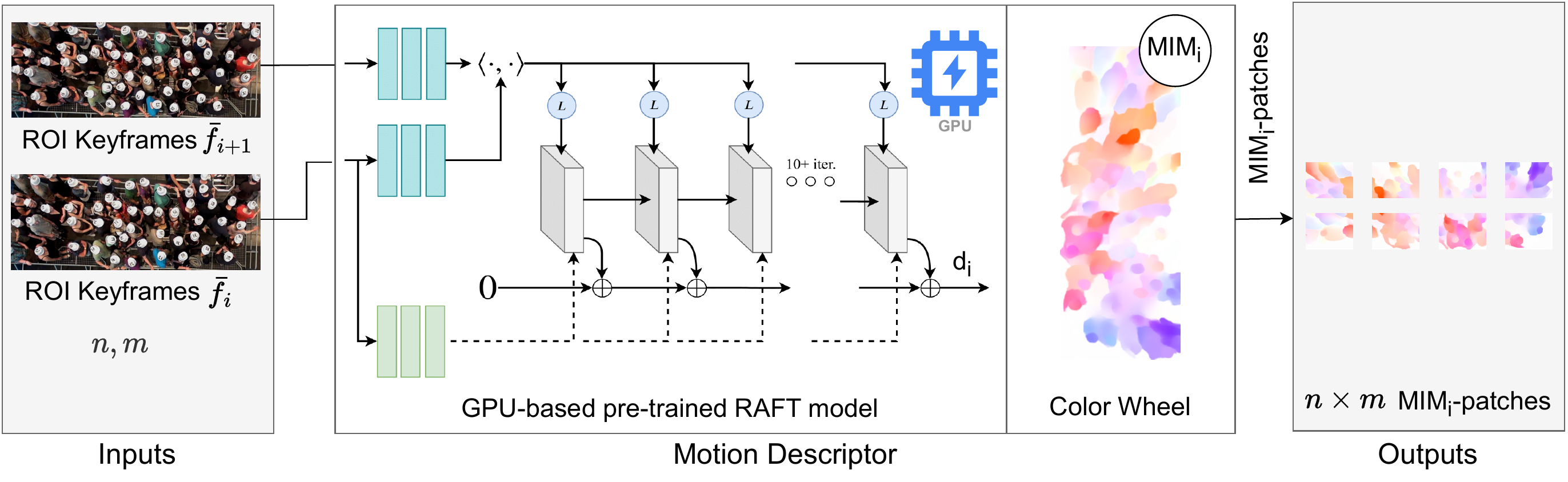}
\caption{Motion descriptor component pipeline. $i$ is the order of the ROI keyframe in the stream. $d$ refers to a dense displacement field. MIM represents motion information map.   } 
\label{fig:MIM-pipeline}
\end{figure*}

\begin{figure}[ht]
\centering
\includegraphics[width=0.95\linewidth]{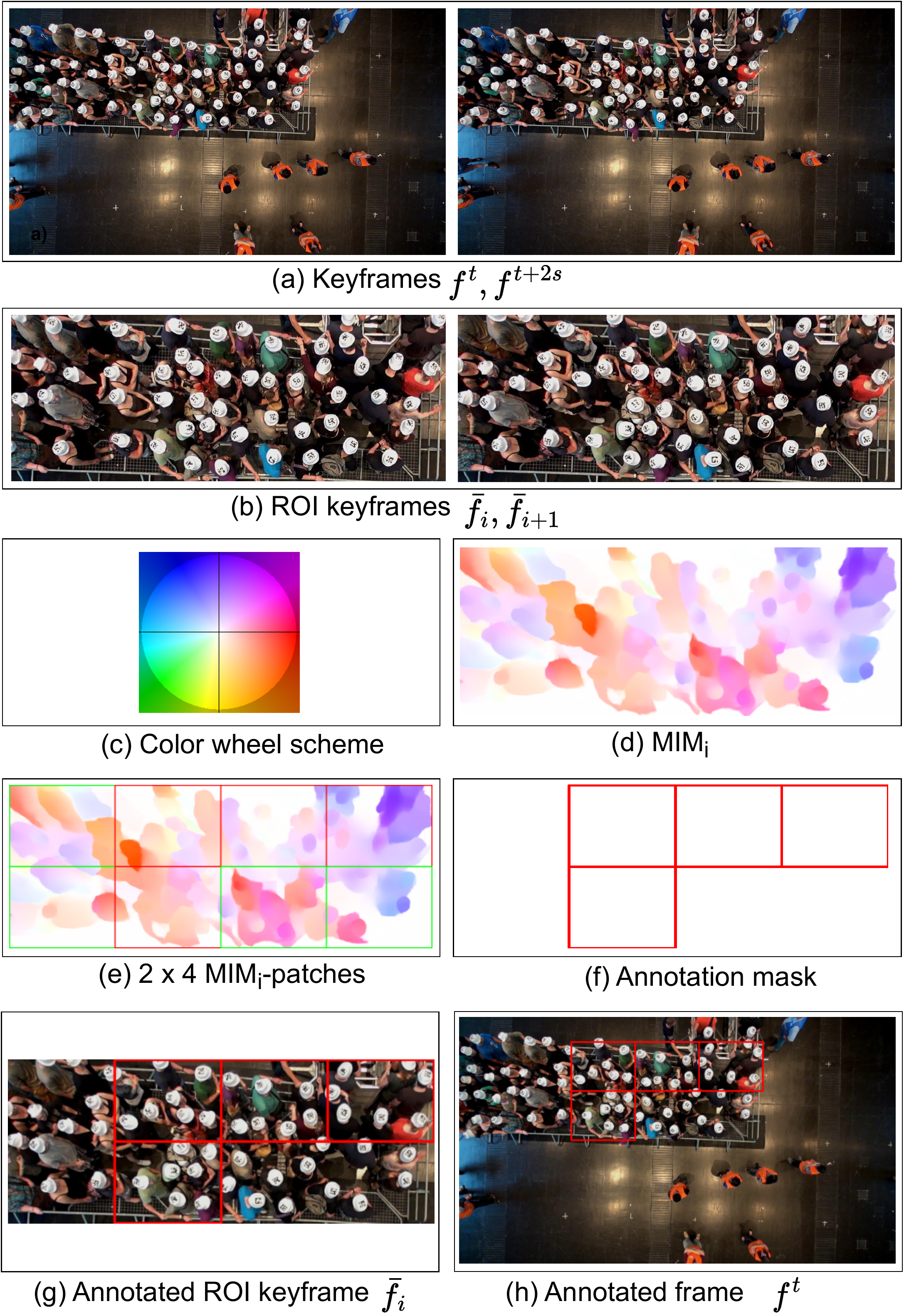}
\caption{An illustration of two keyframes (experiment Entrance\_2 ~\cite{entrance2}, two ROI keyframes,  color wheel schema~\cite{baker2011database}, MIM, $2 \times 4$ MIM-patches, annotation mask, annotated ROI keyframe and annotated frame.  $t$ is the time of the frame $f$ in the stream. $i$ is the order of the ROI keyframe in the stream. $s$  means second. The red boxes indicate pushing patches, while the green boxes mean non-pushing patches.   } 
\label{fig:examples}
\end{figure}

Using this component, we aim to extract the crowd's motion characteristics at the patch level. More specifically, this component estimates the motion direction, magnitude, and associated spatio-temporal information from the crowds, and accordingly visualizes this information. The displayed information includes relevant features that are important for representing the pushing behavior.
As shown in~\cref{fig:MIM-pipeline}, the component uses GPU-based pre-trained RAFT
model and color wheel method to achieve its purpose.
Unlike the majority of the already used optical flow methods~\cite{farneback2003two,yin2019hierarchical},  a GPU-based pre-trained RAFT model performs well in terms of speed, accuracy, and generality for dense crowds~\cite{alia2022hybrid, teed2020raft}.
This model was created by training an ensemble of CNN and recurrent neural networks on the Sintel dataset to calculate the optical flow between two images.
For further details about the model,  we refer the reader to~\cite{teed2020raft}.
Firstly, the component uses the pre-trained model to calculate the displacement of each pixel $\langle x,y  \rangle$ between each pair of  $\bar{f_i}$ and $\bar{f}_{i+1}$, generating the dense displacement field $d_i$. Each pixel location $\langle x,y \rangle$ in $d_i$ is presented by a vector, given by

\begin{equation}
\label{eq:OF_vector}
\textit{\langle u_{\langle x,y\rangle}, v_{\langle x,y\rangle} }\textit{\rangle_{\bar{f_i}, \bar{f}_{i+1}}}  =RAFT(\textit{\langle x,y\rangle_{\bar{f_i}, \bar{f}_{i+1}} } ),
\end{equation}
where $u$ and $v$ are horizontal and vertical displacements of a pixel at the $\langle x,y\rangle$ location between $\bar{f_i}$ and $ \bar{f}_{i+1}$,  respectively. 
This implies that $d_i$ is a matrix of the vectors, as described in
\begin{equation}
\label{eq:dense_OF}
\textit{d_i=\bigg\{ \langle u_{\langle x,y\rangle}, v_{\langle x,y\rangle} \rangle_{\bar{f_i}, \bar{f}_{i+1}} \bigg\} _{(x, y)=(1,1)}^{(w,h)},}
\end{equation}
where $w$ and $h$ are the $\bar{f_i}$ width and height, respectively.

After the estimation of $d_i$, the descriptor applies the color wheel method to deduce the visual motion information from $d_i$. It begins by calculating the direction $\theta$ and magnitude of each vector $\langle u_{\langle x,y\rangle}, v_{\langle x,y\rangle} \rangle$   in $d_i$ using~\cref{eq:angle} and ~\cref{eq:magnitude}, respectively. The color wheel then visualizes the magnitude and direction information to generate MIM$_i$  from the calculated information, where MIM$_i \in \mathbb{R}^{w \times h \times 3}$, and $3$ is the number of channels in MIM$_i$.
~\cref{fig:examples}c is the color wheel scheme,  and~\cref{fig:examples}d is an example of MIM$_i$ that is generated from the pair of  $\bar{f_{i}}$ and $\bar{f}_{i+1}$ (\cref{fig:examples}b). 
According to the wheel schema, the color represents the motion direction, while the color intensity denotes the motion magnitude or speed.
\begin{equation}
\label{eq:angle}
\theta({\textit{\langle x,y \rangle})_{\bar{f_i}, \bar{f}_{i+1}}} =\pi^{-1}\arctan(\frac{\textit{v}_{\langle\textit{ x,y} \rangle}}{\textit{u}_{\langle\textit{ x,y} \rangle}}).
\end{equation}

\begin{equation}
\label{eq:magnitude}
mag({\textit{\langle x,y \rangle}})_{\textit{c_i}} = \sqrt{\textit{u_{\langle x,y \rangle}^2 + v_{\langle x,y \rangle}^2}}
\end{equation}

The motion descriptor component divides each MIM$_i$ into $n \times m$ MIM$_i$-patches to help the framework localizing pushing in ROI.
The MIM$_i$-patches can be expressed as $\{p_{i,k} \in \mathbb{R}^{(w/m) \,  \times\, (h/n)\, \times \,3} \, |\; k=1, 2,\cdots, n\times m\}$,
where  $k$ is the order of the patch in MIM$_i$.
For more clarity,  MIM$_i$ (\cref{fig:examples}d) is divided into $2 \times 4$ MIM-patches (\cref{fig:examples}e). 
It is worth noting that the patch should cover an area on the ground that can accommodate a group of pedestrians, as crowd characteristics are required for representing pushing behavior. To summarize, the MIM-patches represent the output of the motion descriptor component and the input of the next component.

\subsection{Pushing detection and annotation }

The primary purpose of this component (\cref{fig:pushingDetection-pipeline}a) is to localize the pushing patches in the live stream, as well as blurring and storing the annotated ROI keyframes in the cloud storage. Labeling MIM-patches as pushing or non-pushing is the most important aspect of localizing pushing in the live stream. Therefore, we created an efficient binary classifier by adapting and training the EfficientnetV2B0 CNN architecture~\cite{tan2021efficientnetv2}  from scratch, which is then utilized to label the MIM-patches.

\subsubsection{Adapted EfficinetNetV2B0 Architecture}

\begin{figure*}[t]
\centering
\includegraphics[width=0.9\linewidth]{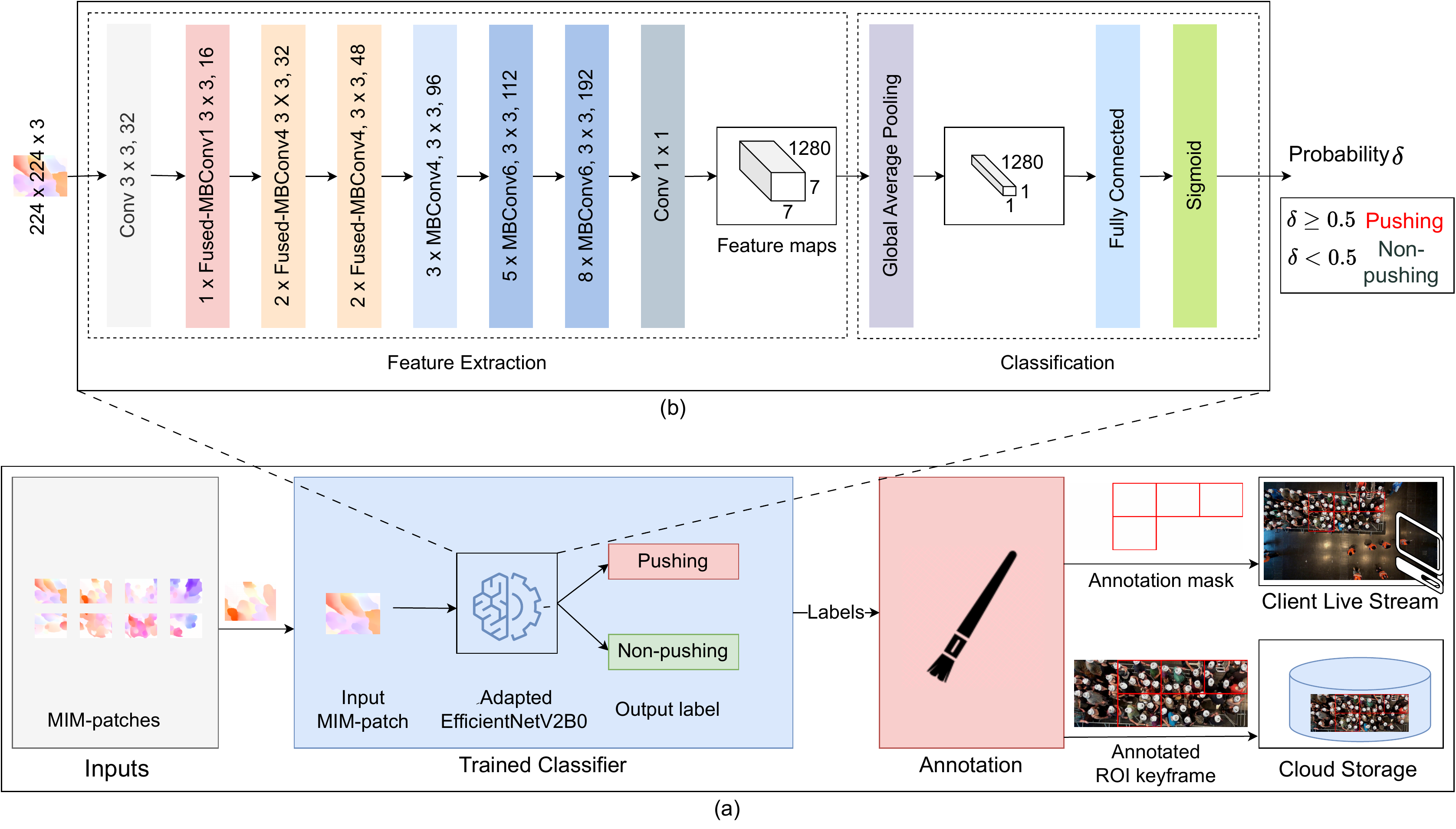}
\caption{(a) The pipeline of pushing detection and annotation component. (b) Adapted EfficientNetV2B0 Architecture.}
\label{fig:pushingDetection-pipeline}
\end{figure*}

EfficientNetV2B0 is a convolutional neural network belonging to the EffivientNetV2 family, designed by the Google Brain team~\cite{tan2021efficientnetv2}. Such a family outperforms state-of-the-art accuracy in different classification tasks with a far smaller model and faster converging speed. EfficientNetV2B0 is the smallest model in this family and achieves high accuracy with minimal computational cost. 

\cref{fig:pushingDetection-pipeline}b depicts the overall architecture of the modified EfficientNetV2B0, which firstly performs a $3 \times 3$ convolution operation on the 
input image, which has dimensions of $224 \times 224 \times 3$. Then it utilizes a combination of 5 Fused-MBConv (Fused Mobile Inverted Residual Bottleneck Convolution)~\cite{gupta2019efficientnet} and 16 MBConv~\cite{sandler2018mobilenetv2} modules for extracting the feature maps (7$\times$7$\times$1280) from the input image. The model then employs a global average pooling layer and a fully connected layer with a Sigmoid activation function for binary classification. The global average pooling2D layer transforms the dimensions of the stacked feature maps to 1$\times$1$\times$1280 and assigns them to the fully connected layer. Finally, the fully connected layer with a Sigmoid activation function finds the probability $\delta$ of the label of the input MIM-patch. Then, the classifier uses the threshold to determine the class of the MIM-patch as~\cref{eq:class}:
\begin{equation}
\label{eq:class}
Class(MIM-patch) =
\begin{cases}
\text{pushing} & \text{if } \delta \geq 0.5
\\
\text{non-pushing} & \text{if } \delta < 0.5
 
\end{cases}
\end{equation}

It's important to note that the classification part of this model differs from the original EfficientNetV2B0, which was designed to classify images into 1,000 categories. However, pushing detection requires labeling the input image into one of two possible classes.

\begin{figure}[ht]
\centering
\includegraphics[width=0.9\linewidth]{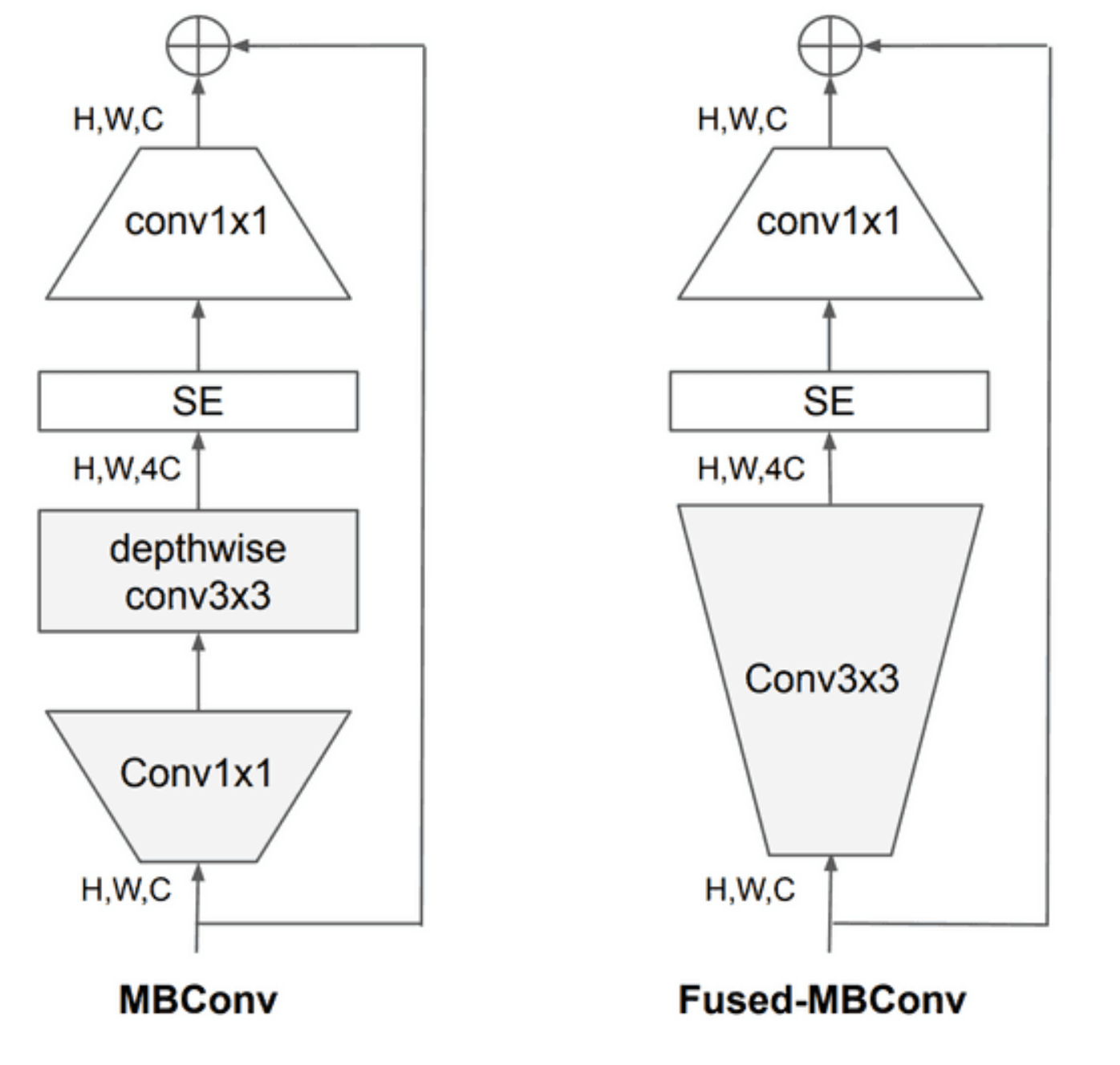}
\caption{Structure of MBConv and Fused-MBConv~\cite{tan2021efficientnetv2}.   } 
\label{fig:MBCon}
\end{figure}

As mentioned above, the main fundamental blocks in EfficientNetV2B0 for feature extraction are MBConv and Fused MBConv~\cite{tan2021efficientnetv2}. As shown in~\cref{fig:MBCon},  MBConv firstly uses a $1\times1$  convolution operation to expand the input activation maps to increase the depth of the feature maps. Next, $3\times3$ depthwise convolutions are applied to reduce the computational complexity and the number of parameters.
Then, a Squeeze-and-Excitation (SE) block enhances the representation power of the architecture.  Finally, another $1\times1$ convolution is employed to reduce the dimensionality of the output feature maps, producing the final output of this block. Moreover, A residual connection is added to enhance the performance further. Despite depthwise convolutions having fewer parameters, they can not often fully utilize modern accelerators. In contrast, the Fused-MBConv tries to solve this problem by replacing the depthwise and expansion conv1$\times$1 in MBConv conv3$\times$3 with a single regular conv3$\times$3, resulting in a faster training process (see~\cref{fig:MBCon}). It is worth mentioning that using only Fused-MBConv in the architecture increases parameters while slowing down the training. Therefore, EfficientNetV2B0 applied a combination of MBConv and Fused-MBConv to improve training speed with a small overhead on parameters and enhance the feature extraction process~\cite{tan2021efficientnetv2}.

The following subsection will discuss the training process for the adapted EfficientNetV2B0 model to classify MIM patches into pushing and non-pushing.

\subsubsection{Adapted EfficientNetV2B0 Training}
\label{sec:trainingprocess}

To classify the MIM-patches into pushing and non-pushing categories, we trained the adapted EfficientNetV2B0 model~(\cref{fig:pushingDetection-pipeline}b) using new training and validation sets comprising both types of MIM-patches (details about the dataset can be found in~\cref{sec:datasetpreparation}). The model parameters used during the training process are listed in~\cref{tab:hyperparameter}, and were chosen based on experimentation to obtain optimal performance with the given dataset. To prevent overfitting, we halted the training if the validation accuracy did not improve after 20 epochs.

\begin{table}[ht]
\centering
\caption{\textbf{The hyperparameter values used in the training process.}}
\label{tab:hyperparameter}
\setlength{\tabcolsep}{3pt}

\begin{tabular}{p{114pt}|p{114pt}}  \hline
Parameter     & Value                \\ \hline
Optimizer     & Adam                 \\
Loss function & Binary cross-entropy \\
Learning rate & 0.001                \\
Batch size    & 32                   \\
Epoch         & 100\\  \hline              
\end{tabular}
\end{table}

~\cref{fig:pushingDetection-pipeline}a shows the pipeline of the pushing detection and annotation component. Firstly, the trained classifier labels   MIM-patches $p_{i,k}$ received from the previous component. Then, the current component displays an annotation mask of the pushing patches in the live stream on the web client. Simultaneously, it blurs and annotates the corresponding ROI keyframe $\bar{f_i}$ before saving it in the cloud storage. Notably, web clients can access this storage via an internet connection.

\section{Evaluation and Results} 
\label{sec:evaluationandresults}

This section introduces the dataset, implementation details, and performance metrics utilized in evaluating the proposed framework. The results of various experiments conducted to assess the performance of our classifier and the proposed framework are also discussed.

\subsection{Dataset Preparation}
\label{sec:datasetpreparation}

Here, we explain how we prepared the labeled dataset (training, validation, and test sets) for training and evaluating the adapted EfficientNetV2B0 as well as all models used in the evaluation. The dataset contains two classes of MIM-patches, which are pushing and non-pushing.

\subsubsection{Data Collection}
In this section, we discuss the data sources used to obtain our dataset. The sources are mainly based on video experiments of crowded event entrances, trajectory data, and ground truth data for pushing behavior. 
Five video experiments with their trajectory data are chosen from the data archive hosted by Forschungszentrum Jülich under CC Attribution 4.0 International license~\cite{crowdqueue, entrance2}.
Static top-view cameras were used to record the videos with a frame rate of 25 frames per second.
It is worth mentioning that the selected experiments contain varied characteristics, which help to improve the generality of the dataset, as seen in~\cref{tab:characterstics}.
The ground truths for the last data source were manually created by social psychologists, who established the definition of pushing behavior in forward motion among crowds~\cite{usten2022pushing}. These ground truths indicate whether the behavior of each pedestrian in every frame is classified as either pushing or non-pushing. 

\begin{figure*}[ht]
\centering
\includegraphics[width=1\linewidth]{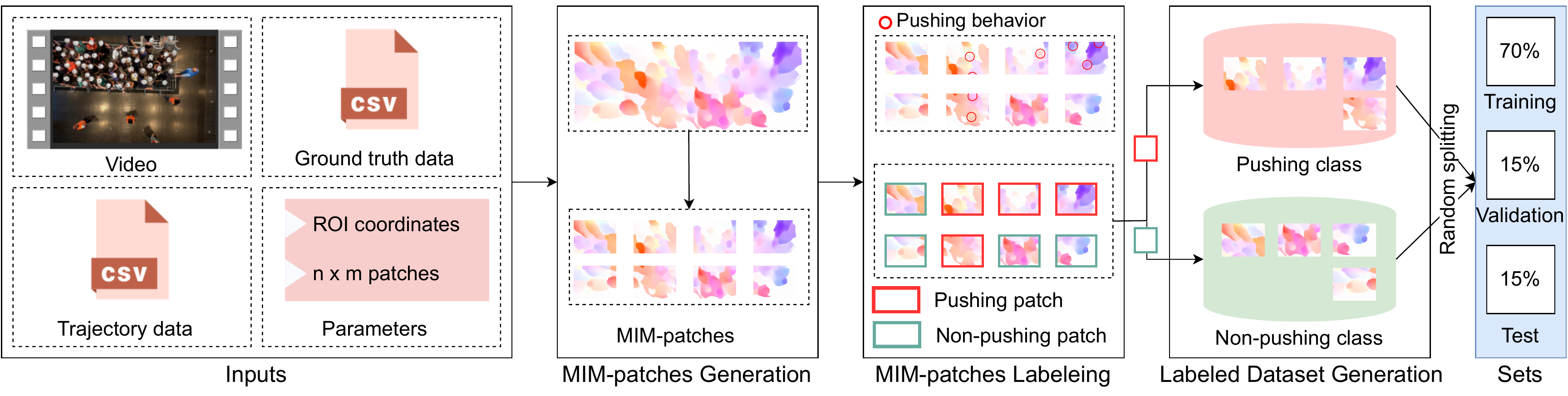}
\caption{The methodology of the labeled dataset generation. }
\label{fig:datasetPreparationMethodology}
\end{figure*}

\begin{table*} 
\centering
\footnotesize
\caption{\label{tab:characterstics}Characteristics of the selected video experiments.  }
\begin{tabular}{l|c|c|c|c|c|c|c|c}
\hline
\textbf{Video}  &
  \textbf{Entrance type} &
 \textbf{Gates} &
  \textbf{Width (m)} &
 \textbf{ Ped.} &
 \textbf{Dur.} &
 \textbf{Resolution} &
 \textbf{ROI coordinates (pixel)} &
\textbf{n $\times$ m patches *} \\ \hline
110         & Straight             & 1 & 1.2 & 63  & 53  & 1920 $\times$  1440 & (374 , 548) , (1382 , 864)  & 1 $\times$ 3 \\  
150         & Straight             & 1 & 5.6 & 57  & 57  & 1920 $\times$ 1440 & (364 , 200) ,  (1378 , 1250)    & 3 $\times$ 3 \\  
270         & Straight             & 1 & 3.4 & 67  & 59  & 1920 $\times$ 1440 & (374 , 330) , (1390 , 1070) & 2 $\times$ 3 \\  
280         & Straight             & 1 & 3.4 & 67  & 67  & 1920 $\times$ 1440 & (374 , 330) , (1390 , 1070) & 2 $\times$ 3 \\  
Entrance\_2 & 90$^{\circ}$  Corner & 2 & 2   & 123 & 125 & 1920 $\times$ 1080 & (213 , 110) , (1337 , 540)  & 2 $\times$ 4\\
 
\hline
\multicolumn{9}{p{510pt}}{The video experiments' names are the same as reported in~\cite{crowdqueue, entrance2}. ``Dur.'' means duration.  ``Ped.'' is an abbreviation for the number of pedestrians. ROI coordinates: left–top and bottom–right coordinates of ROI in the pixel unit. $n \times m$: number of rows
and columns that are used to divide ROI into $n \times m$ regions, which are required for dividing MIM$_i$ into $n \times m$  MIM$_i$-patches. * These values ensure that the dimensions of each region on the ground are greater than one meter,  which is enough to accommodate a group of pedestrian~\cite{alia2022hybrid}. }\\
\end{tabular}

\end{table*}

\subsubsection{Dataset Generation}

The methodology of the labeled dataset generation, as seen in~\cref{fig:datasetPreparationMethodology}, includes three steps: (1)  MIM-patches generation, (2)  MIM-patches labeling and (3) Labeled dataset generation.  

In the MIM-patches generation step,   the motion descriptor component was employed  (\cref{fig:MIM-pipeline}) on the video experiments and their $n \times m$ patches (\cref{tab:characterstics}) to produce  MIM-patches.
To increase the number of patches, the component is applied four times for each video with a different commencement; half a second is the delay duration of each time compared to the previous time. According to~\cite{alia2022hybrid}, half a second delay helps to generate diverse MIM-patches, while less than this period may result in redundant samples.
Based on the trajectory and ground truth data, the second step labels the patches as pushing and non-pushing.
Patches are classified as pushing if it contains at least one pushing behavior, and non-pushing if no pedestrians engage in pushing behavior. On the other hand, the patches that only show a portion of one pedestrian pushing are discarded; because they do not offer complete information about pushing or non-pushing behavior. 
According to the labels of the patches, the last step stores the patches in pushing and non-pushing directories to create the labeled dataset. At the end, the generated dataset consists of 2257 pushing and 1684 non-pushing samples.
To generate the holdout data, the produced dataset is randomly divided into three sets:  \qty{70}{\percent} for training,  \qty{15}{\percent} for validation, and  \qty{15}{\percent} for testing. This split ratio is one of the most commonly used splitting methods in the deep learning field~\cite{genc2019optimal}.
~\cref{tab:sets} shows the number of pushing and non-pushing samples in the training, validation, and test sets.

\begin{table}[h]
\caption{\label{tab:sets} A number of samples in training, validation, and test sets in the generated dataset. }
\begin{tabular}{l|c|c|c|c|c|c|c} \hline
\multicolumn{2}{r|}{Video}           & 110 & 150 & 270 & 280 & E\_2 & \cellcolor[HTML]{cccccc} Total          \\ \hline
\multirow{3}{*}{Training}   & P     & 122 & 182 & 215 & 258 & 808         & \cellcolor[HTML]{cccccc}1585 \\
                            & NP    & 72  & 206 & 197 & 182 & 525         & \cellcolor[HTML]{cccccc}1182 \\
                            & Total & 194 & 388 & 412 & 440 & 1333        & \cellcolor[HTML]{cccccc}2767 \\ \hline
\multirow{3}{*}{Validation} & P     & 26  & 38  & 45  & 55  & 172         & \cellcolor[HTML]{cccccc}336  \\
                            & NP    & 15  & 44  & 42  & 38  & 112         & \cellcolor[HTML]{cccccc}251  \\
                            & Total & 41  & 82  & 87  & 93  & 284         & \cellcolor[HTML]{cccccc}587  \\ \hline
\multirow{3}{*}{Test}       & P     & 26  & 38  & 45  & 55  & 172         & \cellcolor[HTML]{cccccc}336  \\
                            & NP    & 15  & 44  & 42  & 38  & 112         & \cellcolor[HTML]{cccccc}251  \\
                            & Total & 41  & 82  & 87  & 93  & 284         & \cellcolor[HTML]{cccccc}587  \\ \hline
All                     & Total & 276 & 552 & 586 & 626 & 1901        & \cellcolor[HTML]{cccccc}3941 \\ \hline
\multicolumn{8}{p{240pt}}{``All'' refers to all sets. P means pushing. NP is non-pushing. }
\end{tabular}
\end{table}

\subsection{Implementation Details and Evaluation Metrics}


In this article, all the experiments and implementations were conducted on Google Colaboratory Pro (with a GPU NVIDIA of 15 GB and system RAM of 12.7 GB), utilizing JavaScript and Python 3 programming languages along with Keras, TensorFlow 2.0, and OpenCV libraries. Furthermore, all models in the experiments were trained using the same hyperparameter values utilized in the training of our adapted version of EfficientNetV2B0 (\cref{tab:hyperparameter}).

In order to evaluate the performance of our framework,  we utilized a combination of metrics, including accuracy, macro F1-score, and area under the receiver operating characteristic curve (AUC) over the test set. This set of metrics was necessary due to the imbalanced nature of our dataset~\cite{devries2021using}. In addition to these metrics, computational time was also measured as an essential performance metric. The following  provides a detailed explanation of these metrics.

Accuracy: the ratio of successfully classified  MIM-patches to the total number of samples in the test set, and mathematically can be defined as
\begin{equation}
\label{eq:accuracy}
accuracy=\frac{TP+TN}{TP+FP+TN+FN},
\end{equation}
where TP and TN denote correctly classified pushing and non-pushing patches, respectively. FP and FN represent incorrectly predicted pushing (P) and non-pushing (NP) samples.
Accuracy is not enough to evaluate the classifier's performance over an imbalanced dataset, such as our used dataset. Therefore, we used the macro F1-score and AUC metrics, which are valuable for evaluating imbalanced classification problems. 

Macro F1-score: the mean of class-wise F1-scores as described in the formula below:

\begin{equation}
Macro \; F1-score=\frac{F1-score(P) + F1-score(NP)}{2},
\end{equation}
where F1-score is the harmonic average of precision and recall as described in:


\begin{equation}
\label{eq:f1score}
F1-score=\frac{2 \times precision \times recall}{precision + recall},
\end{equation}
 where recall of pushing class is the ratio of correctly classified pushing  MIM-patches to all pushing samples, while precision of pushing class is the ratio of correctly classified pushing patches out of all the samples labeled as pushing by the classifier.  Recall and precision are defined in~\cref{eq:recall} and~\cref{eq:precision}, respectively.
 
 \begin{equation}
 \label{eq:recall}
recall=\frac{TP}{TP+FN},
\end{equation}

 \begin{equation}
 \label{eq:precision}
precision=\frac{TP}{TP+FP}.
\end{equation}

AUC is  the area under the Receiver Operating Characteristics (ROC) curve. ROC is a graph showing the performance of a classification model at all thresholds. The ROC curve plots the false positive rate on the horizontal axis and the true positive rate on the vertical axis. The AUC value ranges from 0 to 1, while a model with an AUC of 1 is considered perfect, while a value of 0.5 indicates that the model performs no better than random guessing.

Computational time: this metric was employed to calculate how long the proposed framework takes to read, analyze and annotate every input, which is two seconds of stream. In other words, computational time determines whether our framework can detect pushing patches within a reasonable time or not.

\subsection{Evaluation of Our Classifier Performance }

We conducted three main comparative empirical experiments to evaluate the effect of our modified EfficientNetV2B0 classifier on the performance of the proposed framework. The first experiment compared
 the proposed classifier against eleven of the most popular CNN architectures. In the second experiment, we compared it to two custom CNN architectures designed for detecting abnormal behavior. Lastly, it was compared to CNN architecture used for pushing detection. Our classifier and all other models were implemented, trained, and assessed utilizing the same MIM-patches dataset, environment, and settings. Moreover, we utilized accuracy, F1-score, and AUC  metrics to measure each model's performance.

\subsubsection{A Comparison with Eleven Popular CNNs }
\cref{tab:popularCNNresults}  depicts the popular CNN architectures used in the first experiment, as well as the comparison results. It is clear that the adapted version of the EfficientNetV2B0 classifier outperformed the rest of the exploited classifiers. In particular, the proposed classifier achieved  \qty{87}{\percent} accuracy and \qty{86}{\percent} F1-score, whereas the second top model in this comparison, DenseNet169, produced  an \qty{83}{\percent} level of both accuracy and F1-score.  This finding is primarily attributable to EfficientNetV2B0's superior efficiency for feature extraction compared to earlier CNN architectures. The main reason for this efficiency is the combination of  MBConv and Fused-MBConv blocks used in EFficientNetV2B0.

\begin{table}[h]
\caption{\label{tab:popularCNNresults} Comparison results to the well-known CNN-based classifiers.}
\centering
\begin{tabular}{l|c|c|c|c}
\hline
  CNN &  Acc. \qty{}{\percent}   & Pre. \qty{}{\percent} & Rec. \qty{}{\percent}  &  F1.  \qty{}{\percent}  \\ \hline
Xception~\cite{chollet2017xception}             & 81 & 81 & 81         & 81          \\ \hline
 VGG16~\cite{simonyan2014very}                & 57  & 36 & 29        & 50         \\ \hline
 VGG19~\cite{simonyan2014very}                & 61  & 61 &62        & 62          \\ \hline
 ResNet50~\cite{he2016deep}             & 80   & 79 & 81        & 79         \\ \hline
 ResNet50V2~\cite{he2016identity}           & 77  & 76 & 77        & 75          \\ \hline
 ResNet101~\cite{he2016deep}            & 72    & 70 & 72      & 70          \\ \hline
 ResNet101V2~\cite{he2016identity}          & 72   & 72 & 72       & 71          \\ \hline
 ResNet152V2~\cite{he2016identity}           & 74   & 73 & 73       & 73          \\ \hline
 
 DenseNet121~\cite{huang2017densely}          & 79     & 79 & 79     & 79          \\ \hline
 DenseNet169~\cite{huang2017densely}          & 83    & 83 & 83      & 83          \\ \hline
 NASNetMobile~\cite{zoph2018learning}         & 57    & 56 & 56      & 56 
\\ \hline \rowcolor{lightgray}

 \textbf{Our classifier }        & \textbf{87} & 87 & 86         & \textbf{86}  \\ \hline
\multicolumn{5}{p{230pt}}{``Acc.'' refers to Accuracy. ``Pre.'' stands for Precision. ``Rec.'' means Recall. ``F1.'' means F1-score. }
 
\end{tabular}
\end{table}

\begin{figure}[ht]
\centering
\includegraphics[width=1\linewidth]{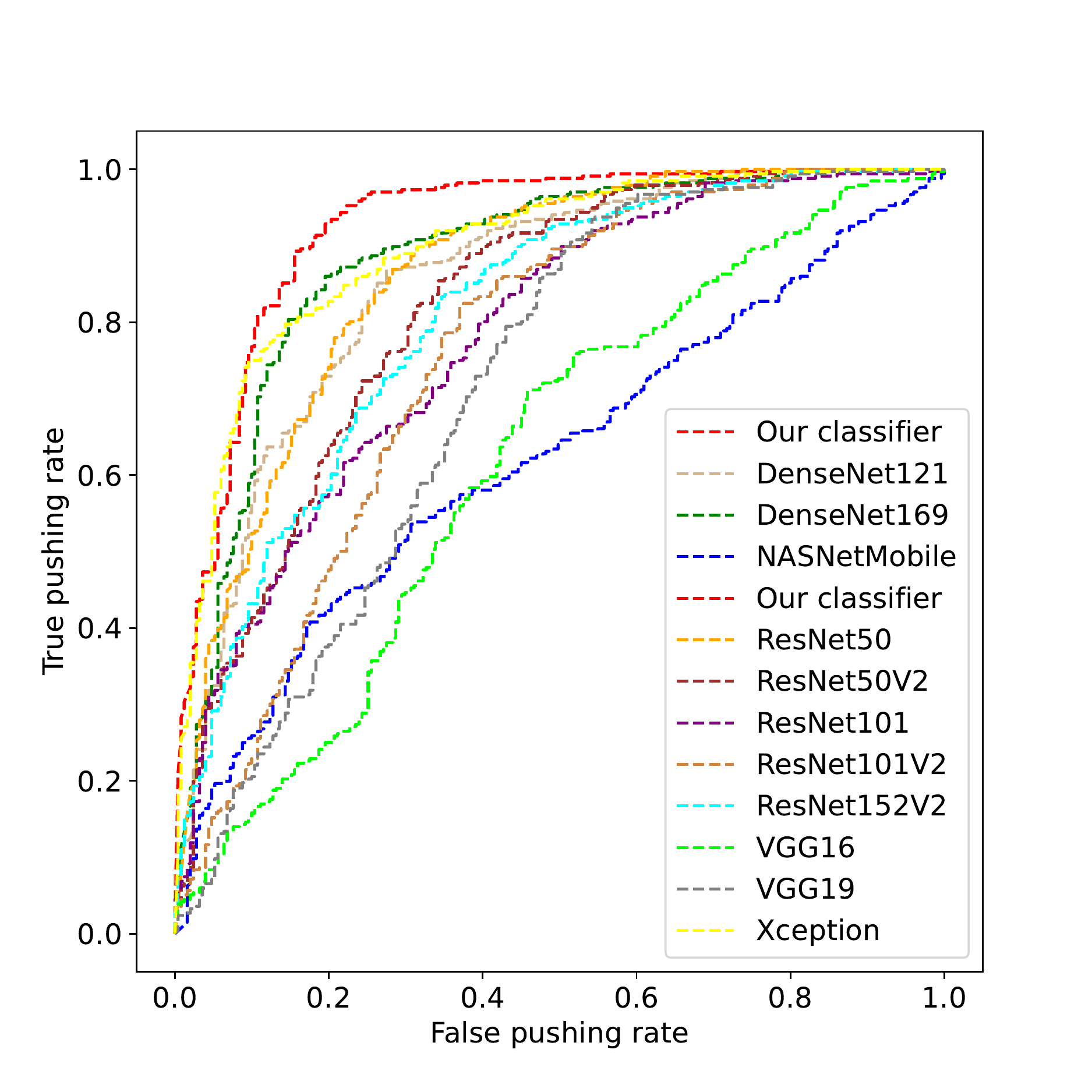}
\caption{ROC curves of our classifier and eleven popular CNN models.   } 
\label{fig:aucpopularcnns}
\end{figure}

Furthermore, as shown in~\cref{fig:aucpopularcnns}, the proposed classifier obtained the highest AUC score (\qty{93}{\percent}) among all the models tested, while the next best model achieving \qty{85}{\percent}. 

\subsubsection{A Comparison with Customized CNNs in Abnormal Behavior Detection}

Here, we have two objectives, 1) Evaluating the performance of some existing CNN models developed to detect abnormal human behavior for pushing detection purposes. 2) Further evaluation of our classifier. The customized architectures are CNN-1~\cite{direkoglu2020abnormal} and CNN-2~\cite{tay2019robust}. The first architecture, CNN-1, employed \qtyproduct{75 x 75}{} pixels as an input image. Furthermore, three convolutional layers, batch normalization, and max pooling operations were used for feature extraction. The developers of this model utilized a fully connected layer with a softmax activation function for classification. The second architecture, CNN-2, downsized the input images to   \qtyproduct{32 x 32}{}  pixels before employing three convolutional layers with three max-pooling layers. For classification, it used two fully connected layers, with the first layer based on a ReLU activation function and the second layer employing a softmax activation function.  

\begin{figure}[ht]
\centering
\includegraphics[width=1\linewidth]{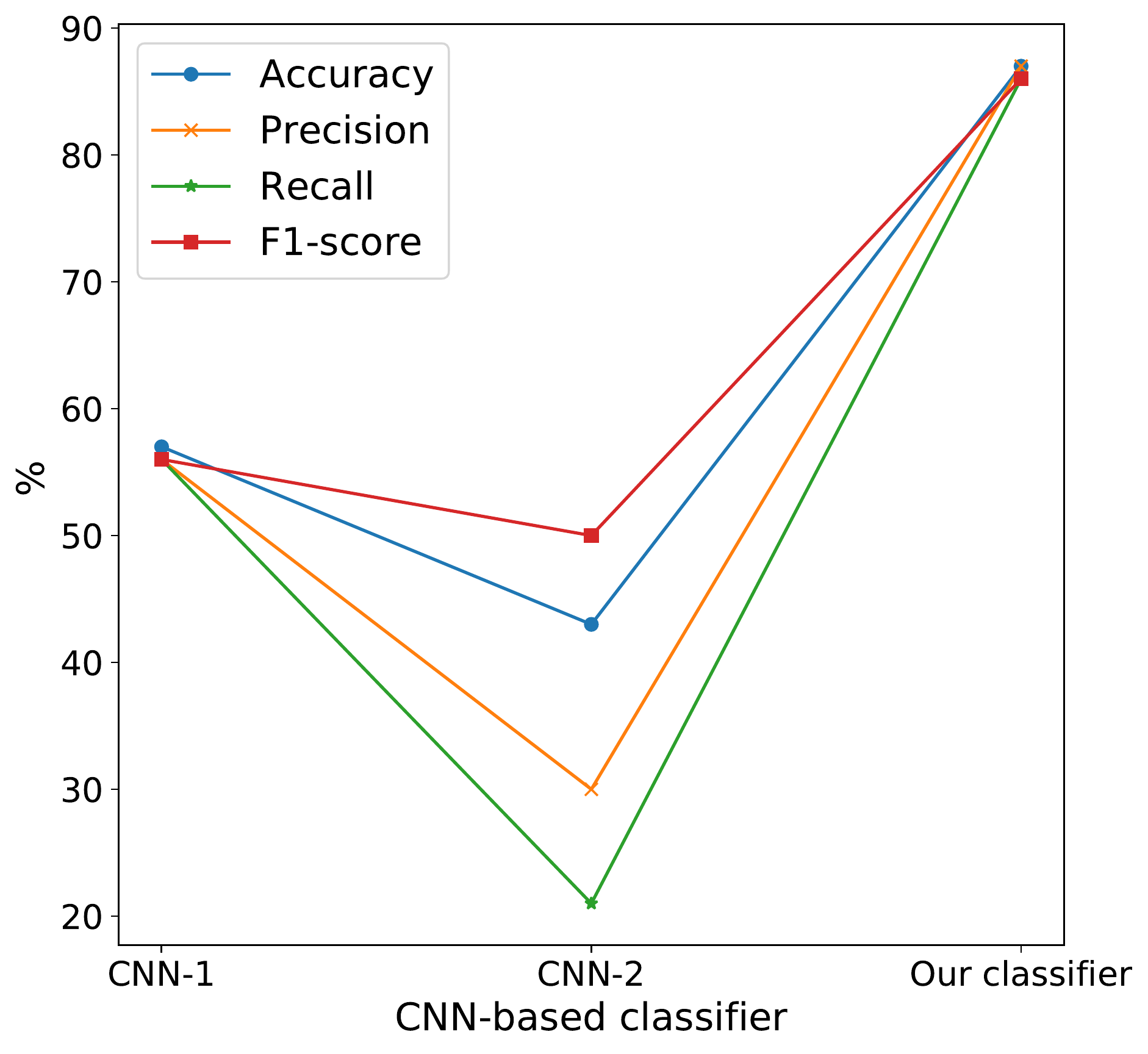}
\caption{Comparison results of our classifier and two customized CNNs.   } 
\label{fig:cuscnns}
\end{figure}

\begin{figure}[ht]
\centering
\includegraphics[width=1\linewidth]{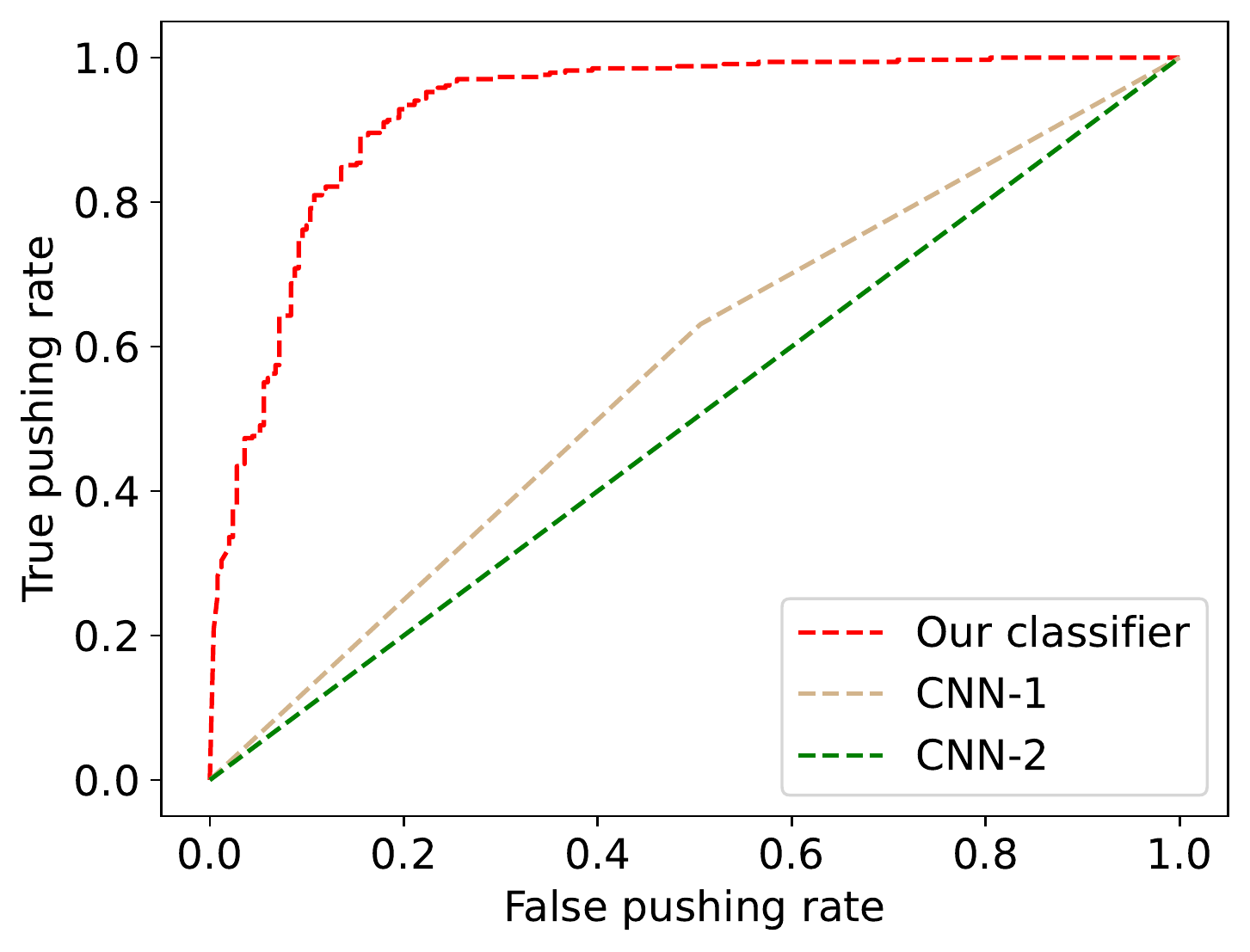}
\caption{ROC curves of our classifier and the two customized CNNs.  } 
\label{fig:auccustomized}
\end{figure}

The results in~\cref{fig:cuscnns} and \cref{fig:auccustomized} show that our classifier surpassed the two classifiers in terms of accuracy,  F1-score and AUC. Furthermore, as pushing detection in crowded scenarios is highly complex, CNN-1 and CNN-2's simple architectures failed to identify pushing MIM-patches. In particular, CNN-1 outperformed CNN-2, but it still produced unsatisfactory outcomes with accuracy, F1-score, and AUC values of \qty{57}{\percent} , \qty{56}{\percent}, and \qty{56}{\percent}, respectively.

\subsubsection{A Comparison with Related Work in Pushing Detection }

\begin{figure}[ht]
\centering
\includegraphics[width=1\linewidth]{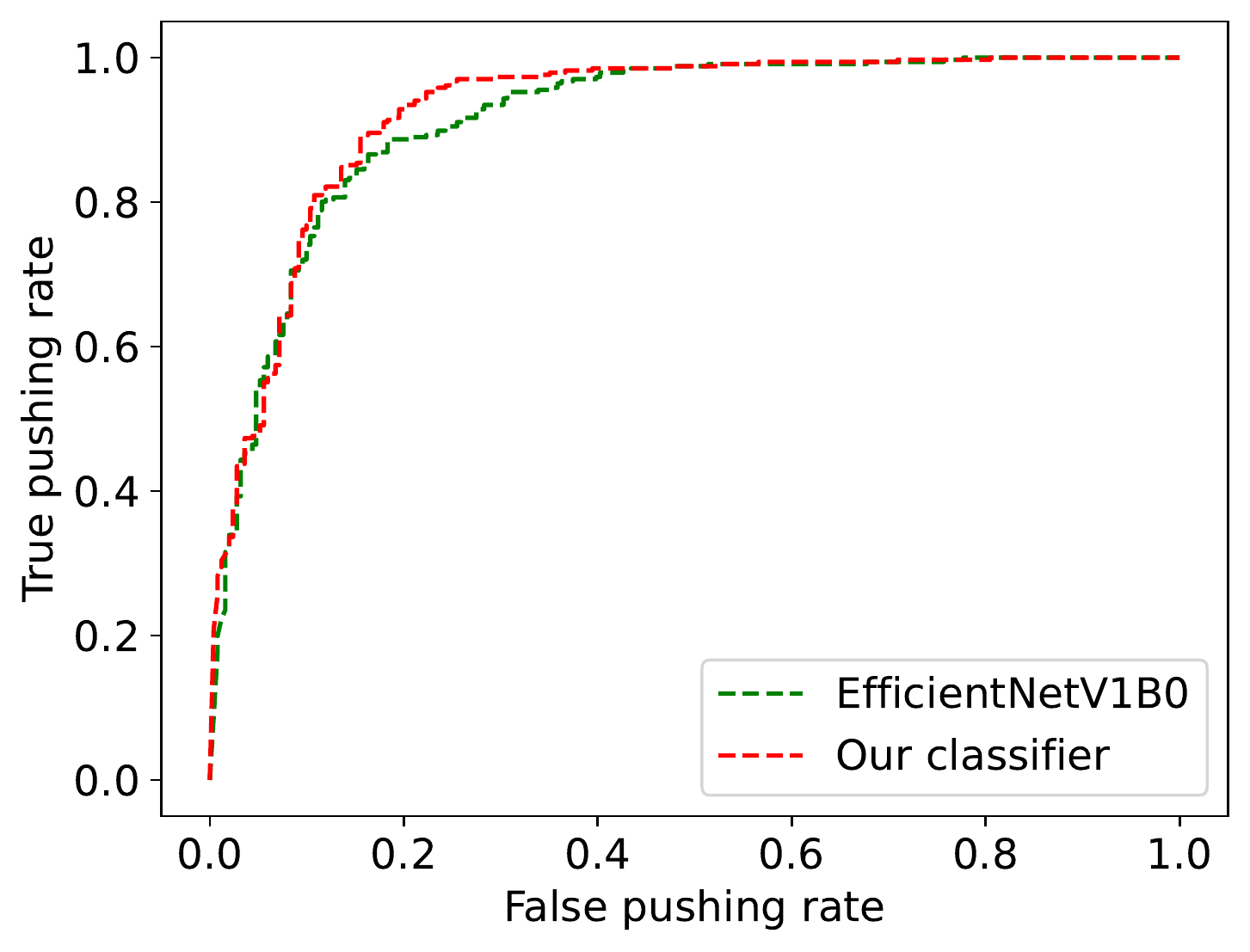}
\caption{ROC curves of our classifier and EfficientNetV1B0. } 
\label{fig:auc-sor}
\end{figure}

Here, we compare the proposed classifier with the CNN architecture (EfficientNetV1B0) employed in Ref.~\cite{alia2022hybrid}, which is the only published work for detecting pushing behavior for forward motion.  Notably, this work does not meet the early identification requirements. As demonstrated in~\cref{tab:state-of-art-results} and \cref{fig:auc-sor}, our combination of adapted EfficientNetV2B0 and MIMs achieved better performance than integrating EfficientNetV1B0 with MIMs by a margin of at least \qty{3}{\percent} in accuracy and F1-score. While EfficientNetV1B0 achieved \qty{91}{\percent} AUC, our classifier achieved \qty{93}{\percent}. This comparison highlights that our hybrid approach surpassed the state-of-art method in pushing detection regarding the accuracy, F1-score, and AUC metrics. In~\cref{sec:cpmputataionaltime}, we will analyze the computational time of both approaches.

\begin{table}[h]
\caption{\label{tab:state-of-art-results} Comparison results to the state-of-art pushing detection approach.}
\centering
\begin{tabular}{l|c|c|c|c}
\hline
  CNN &  Acc. \qty{}{\percent}   & Pre. \qty{}{\percent} & Rec. \qty{}{\percent}  &  F1.  \qty{}{\percent}  \\ \hline
State-of-art approach~\cite{alia2022hybrid}             & 83 & 83 & 84         & 83          \\ \hline \rowcolor{lightgray}
 \textbf{Our hybrid approach }        & \textbf{87} & 86 & 87         & \textbf{86}  \\ \hline
\multicolumn{5}{p{230pt}}{``Acc.'' refers to Accuracy. ``Pre.'' stands for Precision. ``Rec.'' means Recall. ``F1.'' means F1-score. }
\end{tabular}
\end{table}

To summarize the three comparisons, the new hybrid approach based on adapted EfficientNetV2B0 and MIMs outperformed all other tested combinations of CNN models and MIMs in the experiments. This superiority is due to the power of MBConv and Fused-MBConv blocks used in EfficientNetV2B0 for learning the features. Based on the experiments, it can be concluded that our classifier enhanced the performance of the proposed framework.  

\subsection{The Overall Framework Evaluation}

To evaluate the quality of the proposed framework,  we not only evaluated its accuracy and F1-score, but also measured the computational time required for each framework component.

\subsubsection{Performance in terms of Accuracy and F1-score }

\begin{figure} 
\centering
\includegraphics[width=0.9\linewidth]{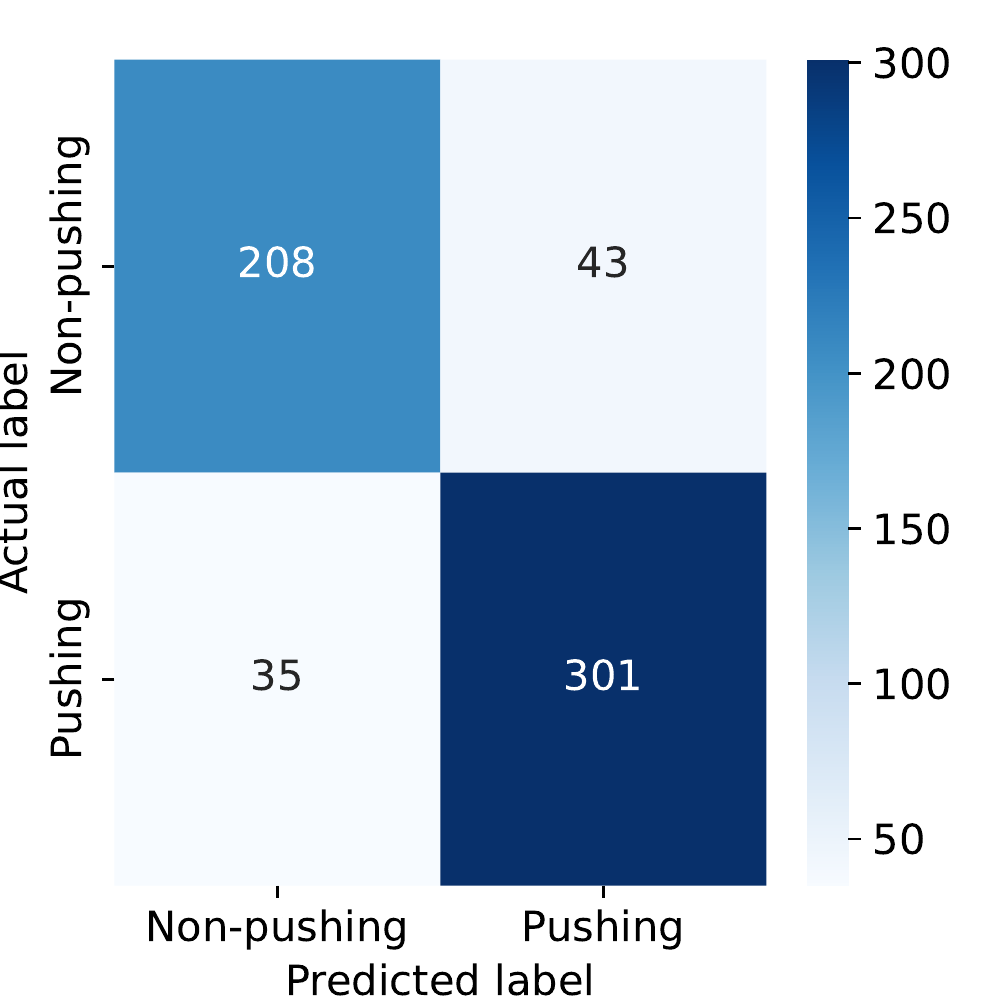}
\caption{Confusion matrix for the proposed framework on all videos.   } 
\label{fig:confusionmatrix}
\end{figure}

The evaluation methodology used comprises several steps as follows:  1) To simulate acquiring the actual inputs, we created a live video stream of crowded event entrances using video recordings of entrances (\cref{tab:characterstics}) and a virtual camera on a web client. In this context, we changed the camera's input to the video recordings. Moreover, we down-scaled the dimensions of each video to half their original resolution to reduce the computational time of the framework.
2) We executed the cloud-based framework to display the live camera stream on the web client, detect pushing patches and record the predicted labels for the test patches in a file.
3) We counted the number of true pushing, false pushing, true non-pushing, and false non-pushing for all videos by comparing the ground truth data with the predicted labels for the test patches,~\cref{fig:confusionmatrix} exhibits the confusion matrix that presents them.
4) Finally, we computed the accuracy and F1-score metrics.
After computing the accuracy and F1-score metrics from the values in the confusion matrix (as shown in~\cref{fig:confusionmatrix}),  our proposed framework achieved an accuracy of \qty{87}{\percent}, precision of \qty{87}{\percent}, recall of \qty{86}{\percent}, and F1-score of \qty{86}{\percent}. These results are consistent with the corresponding quantitative outcomes in our adapted EfficientNetV2B0 classifier over the test set.
 
\subsubsection{Computational Time Analysis}
\label{sec:cpmputataionaltime}

\begin{figure}[htb]
\centering
\includegraphics[width=1\linewidth]{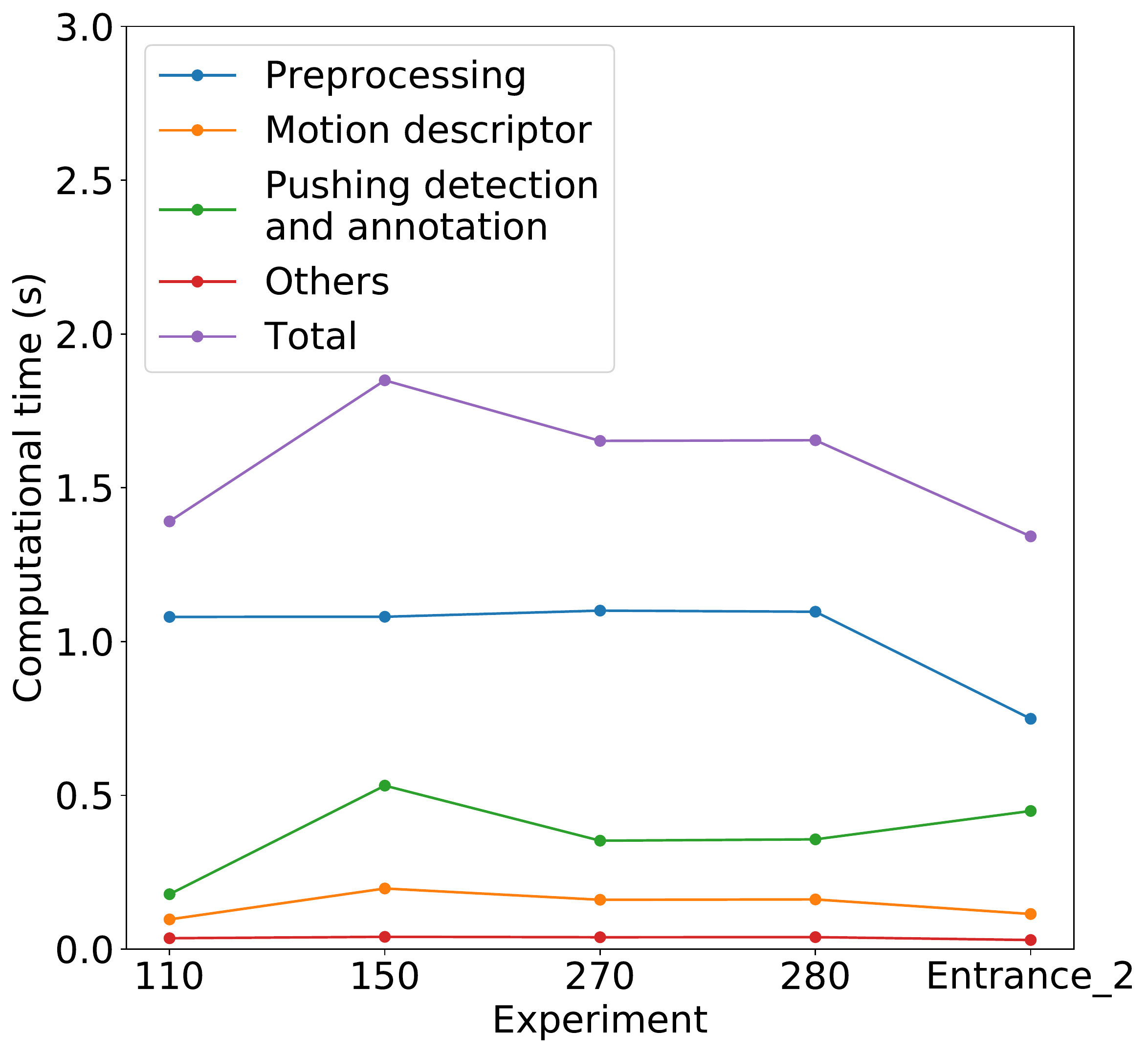}
\caption{Computational time of each component of the proposed framework for annotating two seconds of stream. }
\label{fig:timeOurModel}
\end{figure}

\begin{figure}[htb]
\centering
\includegraphics[width=1\linewidth]{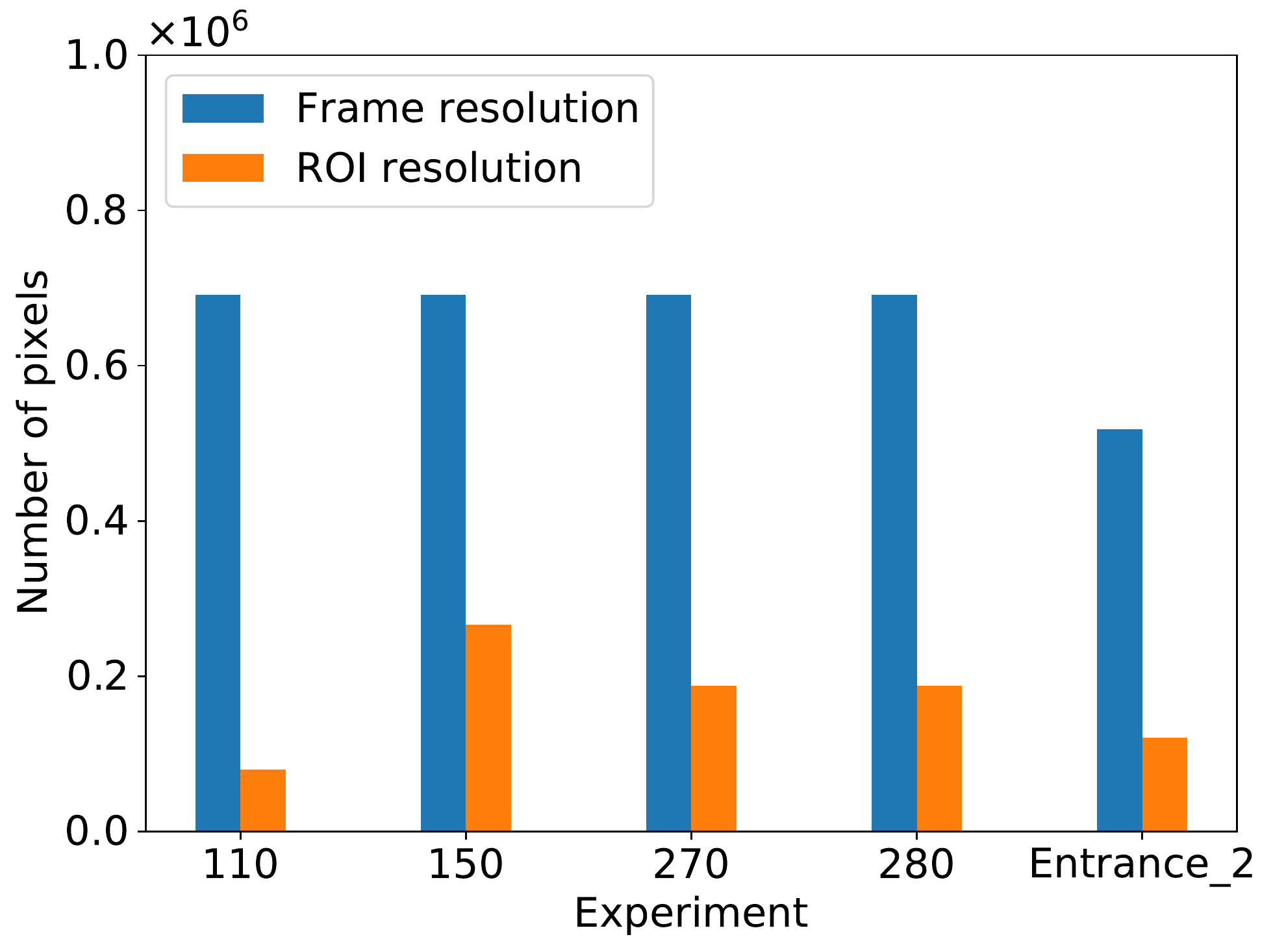}
\caption{Comparison between frame and ROI resolutions for each experiment.}
\label{fig:frameAndROIAreas}
\end{figure}

In order to evaluate the overall computational time of the proposed framework, we computed the required time for each component in the framework. Then, we compared the results against the corresponding parts in the baseline framework~\cite{alia2022hybrid}.
After running both the proposed and baseline frameworks in the same environment using twenty inputs, where each input is a two-second video stream, we calculated the average run-time of all runs. 
As mentioned in the previous paragraph, our framework read the videos using a live camera, while the baseline framework read the same videos directly. \cref{fig:timeOurModel} depicts the time of each component in the proposed framework over every experiment.
As indicated by the produced results, the preprocessing component took more than \qty{50}{\percent} of the overall time to collect and process keyframes from the client camera stream. The resolution of the keyframes primarily determines the time required for this component. For example, experiments 110, 150, 270, and 280 took roughly the same time because they have the same resolution, while experiment entrance\_2  needed less time because its resolution is lower.
In contrast, the motion descriptor component took the least time compared to others, where the ROI resolution plays the most critical role in this component speed. While in the pushing detection and annotation component, the number of patches affects the computational time of this component because each patch requires one classification process.
\cref{fig:frameAndROIAreas} and~\cref{tab:characterstics} display the ROI resolution and the number of patches in each experiment, respectively. 

In general, the computation time increases as the number of patches, frames resolution, and ROI size increase. \cref{fig:timeOurModel} shows that our framework needed less than two seconds to collect, process, detect and annotate each input from the live stream camera.
This means that our framework can annotate the live camera stream within 4 seconds; two seconds for the input duration and lower than two seconds for identifying the pushing patches.

\begin{table*}[!h]
\footnotesize
\centering
\begin{tabular}{l|c|c|c|c} \hline
            & \multicolumn{2}{c|}{\textbf{Baseline framework}} & \multicolumn{2}{c}{\textbf{Our framework}} \\  
\textbf{Experiment} & \multicolumn{1}{l|}{\textbf{Motion information extraction (s)}} & \textbf{Detection (s)} & \textbf{Motion descriptor (s)} & \textbf{Detection and annotation (s)} \\  \hline
110         & 19.42   & 0.15                       & \cellcolor[HTML]{cccccc}0.10           & 0.19          \\  \hline
150         & 19.86     &0.47   &\cellcolor[HTML]{cccccc} 0.20           & 0.53          \\  \hline
270         & 19.38   & 0.32                       & \cellcolor[HTML]{cccccc}0.16           & 0.35          \\  \hline
280         & 19.30  & 0.30                     & \cellcolor[HTML]{cccccc}0.16            & 0.36          \\  \hline
Entrance\_2 & 13.51   & 0.41                     &\cellcolor[HTML]{cccccc} 0.11           & 0.45 \\  \hline        
\end{tabular}
\caption{\label{tab:baselinesystemtime} The computational time of motion descriptor and detection components in our and the baseline frameworks.}
\end{table*}

The results in~\cref{tab:baselinesystemtime} show the comparisons between the motion descriptor, and pushing detection and annotation components in our framework with the corresponding parts in the baseline framework. The motion information extraction part in the baseline framework is similar to the motion descriptor component in our framework, whereas the motion information extraction is slow; it needs more than 13.5 seconds to generate  MIM-patches from two seconds of the video stream. The main reason for this slowness is that it employed CPU-based RAFT to estimate the optical flow vectors for all pixels in the frame. To address this problem, the motion descriptor in our framework implemented RAFT on GPU to calculate the optical flow vectors for each pixel in ROIs instead of all pixels in the frame. As shown in~\cref{fig:frameAndROIAreas}, the number of pixels in ROIs is lesser than \qty{40}{\percent} of the total pixels in the corresponding frames. As a result, the new component took 0.2 seconds or less to produce  MIM-patches from the two seconds of the live stream.
On the other hand, the baseline framework's detection part is slightly faster than the detection and annotation component in the proposed framework.
For example, the previous and new components required 0.47 and 0.53 seconds to work with one input from experiment 150, respectively.
It is important to highlight that the detection part in the baseline framework only finds the labels of the patches, whereas the component in our framework labels, annotates, blurs, and stores the inputs.

In summary, the proposed cloud-based framework can annotate the pushing patches in the live camera stream within four seconds and an accuracy rate of \qty{87}{\percent}.

\section{Conclusion}
\label{sec:conclusion}
This paper proposed a novel automatic framework for the early detection of pushing patches in crowded event entrances.
The proposed framework is based on live camera streaming technology, cloud environment, visualization method, and deep learning algorithms.
The framework first displays the live camera stream of the entrances on the web client in real-time.
Then, it relies on the color wheel method and pre-trained RAFT model to extract the visual motion information from the live stream.
After that, the EfficientNetV2B0-based classifier is adapted and trained to identify pushing patches from the extracted information.
Finally, the framework annotates the pushing patches in the live stream on the web client.
Additionally, it stores the annotated data in the cloud storage, where the stored data is blurred to protect people’s privacy.
In order to train and evaluate the classifier, a new dataset was generated using five real-world video experiments and their associated ground truth data.
The experimental results show that the framework identified pushing patches from the live camera stream with  \qty{87}{\percent} accuracy rate within a reasonable time delay.  

One of the current limitations of the proposed framework is that it is only compatible with a fixed and top-view camera. 

In future, the plan is to develop a new pushing data representation method for machine learning. This method aims to generate dynamic patches based on temporal, spatial, and size dimensions, focusing on one pedestrian for labeling each patch.  
This could potentially help to generate a large dataset with a more efficient sample representation.

\section*{Acknowledgment}
The authors are thankful to Anna Sieben, Helena Lügering, and Ezel Üsten for the valuable discussions, manual annotation of
the pushing behavior in the video experiments, and for revising the proposed framework’s output.

\section*{Ethical Approval}
The experiments used in the dataset were conducted according to the guidelines of the Declaration of Helsinki and approved by the ethics board at the University of Wuppertal, Germany. Informed consent was obtained from all subjects involved in the experiments.

\section*{Data and code availability}
All videos and trajectory data used in generating the patch-based dataset were obtained from the data archive hosted by the Forschungszentrum Jülich under CC Attribution 4.0 International license~\cite{entrance2,crowdqueue}. The undistorted video experiments, implementation of the proposed framework, as well as codes used for building and training the models, are publicly available at: \url{https://github.com/PedestrianDynamics/CloudFast-DL4PuDe} (accessed on 15 Jan 2023). The generated patch-based dataset is available from the corresponding authors upon request.

\bibliographystyle{unsrt}
\bibliography{references}

\begin{IEEEbiography}[{\includegraphics[width=1in,height=1.25in,clip,keepaspectratio]{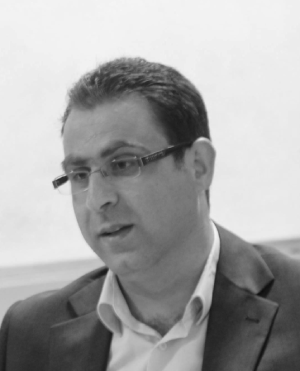}}]{Ahmed Alia} Ahmed Alia is pursuing a Ph.D. in Information Technology at the Institute for Advanced Simulation, Forschungszentrum Jülich, in collaboration with the University of Wuppertal, Germany. He holds a B.Sc. in Computer Science from An-Najah National University, Palestine, and an M.Sc. in Computing from Birzeit University, Palestine. Between 2008 and 2020, he served as a Teaching Assistant, Web Developer, and Trainer at An-Najah National University. His research interests include machine learning, deep learning, computer vision, data analysis, intelligent systems, and crowd behavior analysis.

\end{IEEEbiography}

\vskip 0pt plus -1fil

\begin{IEEEbiography}[{\includegraphics[width=1in,height=1.25in,clip,keepaspectratio]{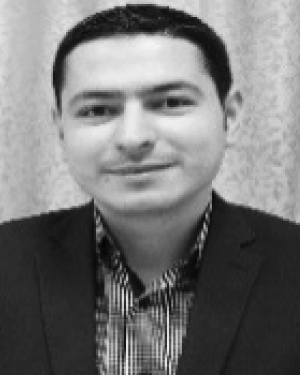}}]{Mohammed Maree}  received the Ph.D. degree in information technology from Monash University. He has published articles in various high-impact journals and conferences, such as ICTAI, Knowledge-Based Systems, Behaviour \& Information Technology, Journal on Computing and Cultural Heritage, Information Development and the Journal of Information Science. He is also a Committee Member/Reviewer of several conferences and journals. He has supervised a number of Master’s and PhD  students in the fields of knowledge engineering, data analysis, information retrieval, natural language processing, and hybrid intelligent systems. He began his career as a Research and Development Manager with gSoft Technology Solution Inc. Then, he worked as the Director of Research and QA with Dimensions Consulting Company. Subsequently, he joined the Faculty of Engineering and Information Technology (EIT), Arab American University, Palestine (AAUP), as a full-time Lecturer. From September 2014 to August 2016, he was the Head of the Multimedia Technology Department, and from September 2016 to August 2018, he was the Head of the Information Technology Department. In addition to his work at AAUP, he worked as a Consultant for SocialDice and Dimensions Consulting companies. Dr. Mohammed is currently an Associate Professor of Information Technology and the Assistant to Vice President for Academic Affairs at the Arab American University.
\end{IEEEbiography}

\vskip 0pt plus -1fil

\begin{IEEEbiography}[{\includegraphics[width=1in,height=1.25in,clip,keepaspectratio]{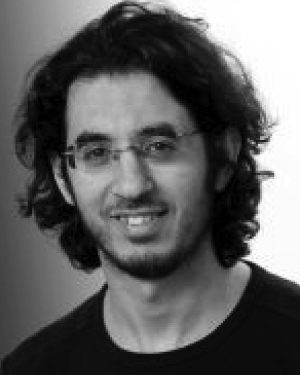}}]{Mohcine Chraibi} received the Diploma degree in computer sciences from the Technical University of Hamburg, Harburg, Germany, and the Ph.D. degree from the Institute for Theoretical Physics, University of Cologne, in 2012, under the supervision of Prof. A. Schadschneider and Prof. A. Seyfried. He worked as a Research Fellow of the JSPS with Tokyo University. Since March 2017, he is the Head of the Pedestrian Dynamics-Modelling Division, Research Centre Jülich, Civil Safety Research Institute. His current researches are focused on the interdisciplinary study of collective dynamics of self-driven particles and its jamming phenomena by means of experiments and simulations.
\end{IEEEbiography}

\begin{IEEEbiography}[{\includegraphics[width=1in,height=1.25in,clip,keepaspectratio]{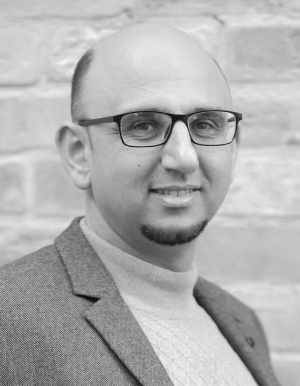}}]{Anas Toma} is an assistant professor at the Department of Computer Engineering, and the coordinator of the Artificial Intelligence Master’s program, An-Najah National University, Palestine. He holds a PhD degree in Computer Engineering from Karlsruhe Institute of Technology (KIT) in Germany. He received his B.Sc. from An-Najah National University in Palestine and his M.Sc. from Jordan University of Science and Technology (JUST) in Jordan. He worked as a teaching and research assistant for one year and as a lecturer for three years at An-Najah National University. From 2016 to 2019, Anas worked as a postdoctoral researcher at TU-Dortmund University in Germany.
\end{IEEEbiography}

\vskip 0pt plus -1fil

\begin{IEEEbiography}[{\includegraphics[width=1in,height=1.25in,clip,keepaspectratio]{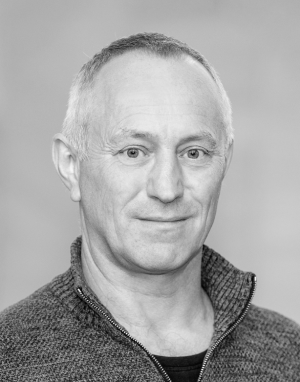}}]{Armin Seyfried }  studied theoretical physics at the Bergische Universität
Wuppertal. For his diploma and doctoral thesis, he focused on
many-particle systems, high-energy physics and parallel computing. After
his doctorate, he was responsible for the modeling and simulation of
building evacuation in an engineering office. From 2004 to 2018 he works
at the Jülich Supercomputing Centre, Forschungszentrum Jülich,
developing models and simulations for application in civil security and
traffic planning. There he established a new research group for
pedestrian and fire dynamics, which became the Institute for Advanced
Simulation-7 in 2018. In addition and since 2010, he is professor for
computer simulations for fire protection and pedestrian traffic at the
University of Wuppertal.
\end{IEEEbiography}

\EOD

\end{document}